\documentclass[10pt,journal,compsoc]{IEEEtran}
\ifCLASSOPTIONcompsoc
  \usepackage[nocompress]{cite}
\else
  \usepackage{cite}
\fi
\ifCLASSINFOpdf
\else
\fi
\usepackage{cite}
\usepackage{amsmath,amssymb,amsfonts}
\usepackage{algorithmic}
\usepackage{graphicx}
\usepackage{textcomp}
\usepackage{xcolor}
\usepackage{epsfig}
\usepackage{epstopdf}
\usepackage{subfigure}
\usepackage{multirow}
\usepackage{tabularx}
\usepackage{comment}
\usepackage{url}
\usepackage{bm}
\usepackage{changepage}
\newcommand{\myfrac}
    [2]{\begin{array}{@{}c@{}}#1 \\[-0.75ex]#2\end{array}}
\newcolumntype{C}[1]{>{\centering\arraybackslash}p{#1}}
\makeatletter
\newcommand{\mathleft}{\@fleqntrue\@mathmargin0pt}
\newcommand{\mathcenter}{\@fleqnfalse}
\makeatother

\begin{document}
%
\title{CHEETAH: An Ultra-Fast, Approximation-Free, and Privacy-Preserved Neural Network Framework based on Joint Obscure Linear and Nonlinear Computations}
%
%
%
%

\author{Qiao~Zhang, Cong~Wang, Chunsheng~Xin,
        and~Hongyi~Wu
\IEEEcompsocitemizethanks{\IEEEcompsocthanksitem Qiao~Zhang, Chunsheng~Xin, and Hongyi~Wu are with the Department
of Electrical and Computer Engineering, Old Dominion University, Norfolk,
VA, 23529. Cong~Wang is with the Department
of Computer Science, Old Dominion University, Norfolk,
VA, 23529.\protect\\
E-mail: \{qzhan002, c1wang, cxin, h1wu\}@odu.edu
}
}

\IEEEtitleabstractindextext{%
\begin{abstract}
Machine Learning as a Service (MLaaS) is enabling a wide range of smart applications on end devices. However, such convenience comes with a cost of privacy because users have to upload their private data to the cloud. This research aims to provide effective and efficient MLaaS such that the cloud server learns nothing about user data and the users cannot infer the proprietary model parameters owned by the server. This work makes the following contributions. First, it unveils the fundamental performance bottleneck of existing schemes due to the heavy permutations in computing linear transformation and the use of communication intensive Garbled Circuits for nonlinear transformation. Second, it introduces an ultra-fast secure MLaaS framework, CHEETAH, which features a carefully crafted secret sharing scheme that runs significantly faster than existing schemes without accuracy loss. Third, CHEETAH is evaluated on the benchmark of well-known, practical deep networks such as AlexNet and VGG-16 on the MNIST and ImageNet datasets. The results demonstrate more than $100\times$ speedup over the fastest GAZELLE (Usenix Security'18), $2000\times$ speedup over MiniONN (ACM CCS'17) and five orders of magnitude speedup over CryptoNets (ICML'16). This significant speedup enables a wide range of practical applications based on privacy-preserved deep neural networks.
\end{abstract}

\begin{IEEEkeywords}
privacy; machine learning as a service; secure two party computation; joint obscure neural computing
\end{IEEEkeywords}}

\maketitle

\IEEEdisplaynontitleabstractindextext

%
\IEEEpeerreviewmaketitle

\IEEEraisesectionheading{\section{Introduction}\label{Intro}}
\IEEEPARstart{F}{rom} Alexa and Google Assistant to self-driving vehicles and Cyborg technologies, deep learning is rapidly advancing and transforming the way we work and live. It is becoming prevalent and pervasive, embedded in many systems, e.g., for pattern recognition~\cite{li2015convolutional}, medical diagnosis \cite{fakoor2013using}, speech recognition~\cite{dahl2012context} and credit-risk assessment~\cite{fan2018denoising}. In particular, deep Convolutional Neural Network (CNN) has demonstrated superior performance in computer vision such as image classification~\cite{krizhevsky2012imagenet,simonyan2014very} and facial recognition~\cite{schroff2015facenet}, among many others.

Since training a deep neural network model is resource-intensive, cloud providers begin to offer Machine Learning as a Service (MLaaS) \cite{wang2018rafiki}, where a proprietary model is trained and hosted on clouds, and clients make queries (inference) and receive  results through a web portal. While this emerging cloud service is embraced as important tools for efficiency and productivity, the interaction between clients and cloud servers creates new vulnerabilities for unauthorized access to private information. This work focuses on ensuring privacy-preserved while efficient inference in MLaaS.

Although communication can be readily secured from end to end, privacy still remains a fundamental challenge. On the one hand,  the clients must submit their data to the cloud for inference, but they want the data privacy well protected, preventing curious cloud provider from mining valuable information. In many domains such as health care~\cite{mozaffari2015systematic} and finance~\cite{sohangir2018big}, data are extremely sensitive. For example, when patients transmit their physiological data to the server for medical diagnosis, they do not want anyone (including the cloud provider) to see it. Regulations such as Health Insurance Portability and Accountability Act (HIPAA) \cite{act1996health} and the recent General Data Protection Regulation (GDPR) in Europe~\cite{file2012proposal} have been in place to impose restrictions on sharing sensitive user information. On the other hand, cloud providers do not want users to be able to extract their proprietary, valuable model that has been trained with significant resource and efforts, as it may turn customers into one-time shoppers~\cite{tramer2016stealing}. Furthermore, the trained model contains private information about the training data set and can be exploited by malicious users~\cite{shokri2017membership,song2017machine,wang2018stealing}. To this end, there is an urgent need to develop effective and efficient schemes to ensure that, in MLaaS, a cloud server does not have access to users' data and a user cannot learn the server's model.

\begin{table*}
\begin{center}
\caption{Comparison of Privacy-Preserved Neural Networks.}
\vspace*{-0.12in}
{\small\begin{tabular}{r||l|l|c}
\hline \hline
\ & Scheme for Linear Computation & Scheme for Non-Linear Computation & Speedup over ~\cite{cryptonets} \\
\hline
CryptoNets~\cite{cryptonets} & HE & HE (square approx.) & --  \\
\hline
Faster CryptoNets~\cite{chou2018faster} & HE & HE (polynomial approx.) & 10$\times$  \\
\hline
GELU-Net~\cite{zhang2018gelu} & HE & Plaintext (no approx.) & 14$\times$  \\
\hline
E2DM~\cite{jiang2018secure} & Packed HE \& Matrix optimization & HE (square approx.) & 30$\times$  \\
\hline
SecureML~\cite{mohassel2017secureml} & HE \& Secret share & GC (piecewise linear approx.) & 60$\times$  \\
\hline
Chameleon \cite{riazi2018chameleon} & Secret share & GMW \& GC (piecewise linear approx.) & 150$\times$  \\
\hline
MiniONN~\cite{liu2017oblivious} & Packed HE \& Secret share & GC (piecewise linear approx.) & 230$\times$  \\
\hline
DeepSecure \cite{rouhani2018deepsecure} & GC & GC (polynomial approx.) & 527$\times$  \\
\hline
SecureNN~\cite{waghsecurenn} & Secret share & GMW (piecewise linear approx.) & 1000$\times$  \\
\hline
FALCON~\cite{li2018falcon} & Packed HE with FFT & GC (piecewise linear approx.) & 1000$\times$  \\
\hline
XONN~\cite{riazi2019xonn} & GC & GC (piecewise linear approx.) & 1000$\times$  \\
\hline
GAZELLE~\cite{juvekar18gazelle} & Packed HE \& Matrix optimization & GC (piecewise linear approx.) & 1000$\times$  \\
\hline
\textbf{CHEETAH}  & \textbf{Packed HE} \& \textbf{Obscure matrix cal.} & \textbf{Obscure HE} \& \textbf{SS (no approx.)} & \textbf{100,000}$\times$  \\
\hline \hline
\end{tabular}}\label{table:comparison}
\end{center}
\vspace*{-0.1in}
\end{table*}
\subsection{Retrospection: Evolvement of Privacy-Preserved Neural Networks}\label{Intro:related}
The quest began in 2016 when CryptoNets~\cite{cryptonets} was proposed to embed Homomorphic Encryption (HE)~\cite{gentry2009fully} into CNN. It was the first work that successfully demonstrated the feasibility of calculating inference over Homomorphically encrypted data. While the idea is conceptually straightforward, its prohibitively high computation cost renders it impractical for most applications that rely on non-trivial deep neural networks with a practical size in order to  characterize complex feature relations~\cite{simonyan2014very}. For instance, CryptoNets takes about $300$s for computing inference even on a simple three-layer CNN architecture. With the increase of depth, the computation time grows exponentially. Moreover, several key functions in neural networks (e.g., activation and pooling) are nonlinear. CryptoNets had to use Taylor approximation, e.g.,  replacing the original activation function with a square function. Such approximation leads to not only degraded accuracy compared with the original model, but also instability and failure in training.

Following CryptoNets, the past two years have seen a multitude of works aiming to improve the computation accuracy and efficiency (as summarized in Table~\ref{table:comparison}). A neural network essentially consists of two types of computations, i.e., linear and nonlinear computations. The former focuses on matrix calculation to compute dot product (for fully-connected dense layers) and convolution (for convolutional layers). The latter includes nonlinear functions such as activation, pooling
and softmax. A series of studies have been carried out to accelerate the linear computation, or nonlinear computation, or both. For example, {\em faster CryptoNets}~\cite{chou2018faster} leveraged sparse polynomial multiplication to accelerate the linear computation. It achieved about 10 times speedup over CryptoNets. SecureML~\cite{mohassel2017secureml}, Chameleon~\cite{riazi2018chameleon} and MiniONN~\cite{liu2017oblivious} adopted a similar design concept. Among them, MiniONN achieved the highest performance gain. It applied \emph{Secret Share} (SS) for linear computation, and packed HE~\cite{fan2012somewhat} to pre-share a noise vector between the client and server offline, in order to cancel the noise during secure online computation. In \cite{liu2017oblivious}, non-linear functions were approximated by piece-wise linear segments, and computed by using Garbled Circuits (GC), which resulted in $230$ times speedup over CryptoNets. DeepSecure~\cite{rouhani2018deepsecure} took an all-GC approach, i.e., implemented both linear and nonlinear computations  using GC. It optimized the gates in the traditional GC module to achieve a speedup of 527 times over CryptoNets. Finally, GAZELLE~\cite{juvekar18gazelle} focused on the linear computation, to accelerate the matrix-vector multiplication based on packed HE, such that Homomorphic computations can be efficiently parallelized on multiple packed ciphertexts. GAZELLE demonstrated impressive speedup of about $20$ times compared with MiniONN and three orders of magnitude faster than CryptoNets. So far, GAZELLE is considered the state-of-art framework for secure inference computation.

Two recent works unofficially published in arXiv reported new designs that achieved computation speed at the same order of magnitude as GAZELLE. FALCON~\cite{li2018falcon} leveraged fast Fourier Transform (FFT) to accelerate linear computation. Its computing speed is similar to GAZELLE, while the communication cost is higher. SecureNN~\cite{waghsecurenn} adopted a design philosophy similar to Chameleon and MiniONN, but exploited
a 3-party setting to accelerate the secure computation, to obtain a 4 times speedup over GAZELLE, at the cost of using a semi-trust third party. Additionally, XONN~\cite{riazi2019xonn} worked in line with DeepSecure to explore the GC based design for Binary Neural Network (BNN), achieving up to 7 times speedup over GAZELLE, at the cost of accuracy drop due to the binary quantization in BNN.

In addition, a few approaches were introduced to not just improve computation efficiency but also provide other desirable features. For example, GELU-Net~\cite{zhang2018gelu} aims to avoid approximation of non-linear functions. It partitioned computation onto non-colluding parties: one party performs linear computations on encrypted data, and the other executes nonpolynomial computation in an unencrypted but secure manner.
It showed over 14 times speedup than CryptoNets and does not have accuracy loss.
E2DM~\cite{jiang2018secure} aimed to encrypt both data and neural network models, assuming the latter are uploaded by users to untrusted cloud. It focused on matrix optimization by combining Homomorphic operation and ciphertext permutation,
demonstrating 30 times speedup over CryptoNets.
\subsection{Contribution of This Work}\label{Intro:contribution}
Despite the fast and promising improvement in computation speed, there is still a significant performance gap to apply privacy-preserved neural networks on practical applications. The time constraints in many real-time applications (such as speech recognition in Alexa and Google Assistant) are within $10$ seconds~\cite{alexainf,googleassis}; self-driving cars even demand an immediate response less than a second~\cite{dixit2016autonomous}. In contrast, our benchmark has showed that GAZELLE, which has achieved the best performance so far in terms of speed among  existing schemes, takes 161s and 1731s to run the well-known practical deep neural networks AlexNet~\cite{krizhevsky2012imagenet} and VGG-16~\cite{simonyan2014very}, which renders it impractical in real-world applications.

In this paper, we propose CHEETAH, an ultra-fast, secure MLaaS framework that features a carefully crafted secret sharing scheme to enable efficient, joint linear and nonlinear computation, so that it can run significantly faster than the state-of-the-art schemes. It eliminates the need to use approximation for nonlinear computations; hence, unlike the existing schemes,  CHEETAH does not have accuracy loss.
It, for the first time, reduces the computation delay to milliseconds and thus enables a wide range of practical applications to utilize  privacy-preserved deep neural networks. To the best of knowledge, this is also the first work that demonstrates privacy-preserved inference based on the well-known, practical deep architectures such as AlexNet and VGG.

The significant performance improvement of CHEETAH stems from a creative design, called {\em joint obscure neural computing}. Computations in neural networks follow a series of operations alternating between  linear and nonlinear transformations for feature extraction. Each operation takes the output from the previous layer as the input. For example, the nonlinear activation is computed on the weighted values of linear transformations (i.e., the dot product or convolution).
All existing approaches discussed in Sec.~\ref{Intro:related} essentially follow the same framework, aiming to securely compute the results for each layer and then propagate to the next layer. This seemingly logic approach, however, becomes the fundamental performance hurdle as revealed by our analysis.

First, although matrix computation has been deeply optimized based on packed HE for the linear transformation in the state-of-the-art GAZELLE, it is still costly. The computation time of the linear transformation is dominated by the operation called ciphertext permutation (or Perm)~\cite{juvekar18gazelle}, which generates the sum based on a packed vector. It is required in both convolution (for a convolutional layer) and dot product (for a dense layer).
From our experiments, one Perm is 56 times slower than one Homomorphic addition and 34 times slower than one Homomorphic multiplication.
We propose an approach to enable an incomplete (or obscure) linear transformation result to propagate to the next nonlinear transformation as the input to continue the neural computation, reducing the number of ciphertext permutations to zero in both convolution and linear dot product computation.

Second, most existing schemes (including GAZELLE) adopted GC to compute the nonlinear transformation (such as activation, pooling
and softmax), because GC generally performs better than HE when the multiplicative depth is greater than 0 (i.e., nonlinear)~\cite{juvekar18gazelle}. However, the GC-based approach is still costly. The overall network must be represented as circuits and involves interactive communications between two parties to jointly evaluate neural functions over their private inputs. The time cost is often significant for large and deep networks.
Specifically, our benchmark shows that it takes about 263 seconds to compute a nonlinear  ReLu function with 3.2M input values, which is part of the VGG-16 framework \cite{simonyan2014very}. Moreover, all existing GC-based solutions rely on piece-wise or polynomial approximation for nonlinear functions such as activation. This leads to degraded accuracy and the accuracy loss is often significant. The proposed scheme takes a secret sharing approach with 0-multiplicative-depth packed HE to avoid the use of computationally expensive GC. A novel design is developed to allow the server and client to each obtain a share of Homomorphic encrypted nonlinear transformation result based on the obscure linear transformation as discussed above. This approach eliminates the need to use approximation for nonlinear functions and achieves enormous speedup.
For example, it is 1793 times faster than GAZELLE in computing the most common nonlinear ReLu activation function, under the output dimension of 10K.

Overall, the proposed CHEETAH is an ultra-fast privacy-preserved neural network inference framework without accuracy loss. It enables obscure neural computing that intrinsically merges the calculation of linear and nonlinear transformations and effectively reduces the computation time. We benchmark the performance of CHEETAH with well-known deep networks for secure inference. Our results show that it is  218 and 334 times faster than GAZELLE, respectively, for a 3-layer and a 4-layer CNN used in previous works. It achieves a significant speedup of 130  and 140 times, respectively, over GAZELLE in the well-known, practical deep networks AlexNet and VGG-16. Compared with CryptoNets, CHEETAH achieves a speedup of five orders of magnitudes.

The rest of the paper is organized as follows. Section \ref{prery} introduces the system and threat models. Section \ref{systm} elaborates the system design of CHEETAH, followed by the security analysis in Section \ref{secr}. Experimental results are discussed in Section \ref{perf}. Finally, Section \ref{conclu} concludes the paper.

\section{System and Threat Models}\label{prery}
In this section, we introduce the overall system architecture and threat model, as well as the background knowledge about cryptographic tools used in our design.
\subsection{System Model}
\begin{figure}[!b]
\vspace*{-0.25in}
\centering
\includegraphics[width=1\columnwidth]{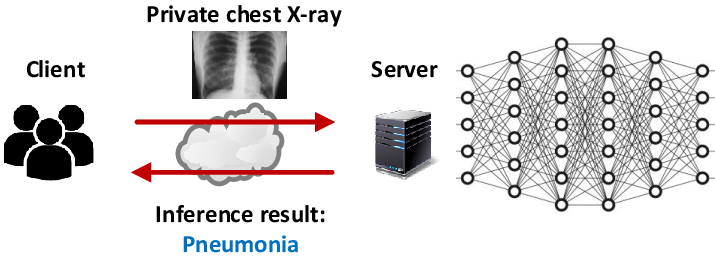}
\caption{An overview of the MLaaS system.}
\label{mlaas}
\end{figure}

We consider a MLaaS system as shown in Fig.~\ref{mlaas}. The client is the party that generates or owns the private data. The server is the party that has a well-trained deep learning model and provides the inference service based on the client's data. For example, a doctor performs a chest X-ray for her patient and sends the X-ray image to the server on the cloud, which runs the neural network model and returns the inference result to assist the doctor's diagnosis. 

\begin{figure}[!b]
\vspace*{-0.17in}
\centering
\subfigure[Overall network structure.] { \label{overview:A}
\includegraphics[width=0.9\columnwidth]{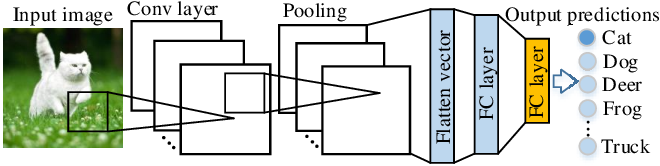}
}
\vspace*{-0.13in}
\subfigure[Convolutional layer.] { \label{overview:a}
\includegraphics[width=0.9\columnwidth]{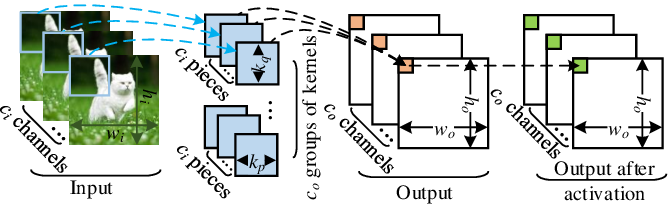}
}

\subfigure[Pooling.] { \label{overview:b}
\includegraphics[width=0.46\columnwidth]{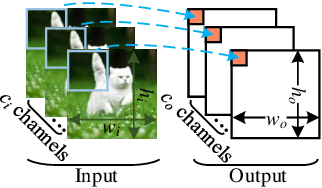}
}
\subfigure[Fully connected layer.] { \label{overview:c}
\includegraphics[width=0.46\columnwidth]{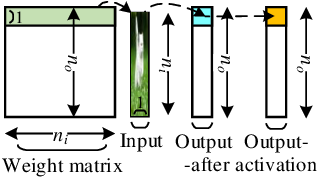}
}
\vspace*{-0.08in}
\caption{A three-layer CNN: (a) overall network structure, (b) convolutional layer, (c) pooling, (d) fully connected layer.}
\label{overview}
\vspace*{-0.03in}
\end{figure}

While various deep learning techniques can be employed to enable MLaaS, we focus on the Convolutional Neural Network (CNN), which has achieved wide success and  demonstrated superior performance in computer vision such as image classification~\cite{krizhevsky2012imagenet,simonyan2014very} and face recognition~\cite{schroff2015facenet}. A CNN consists of a stack of layers to learn a complex relation among the input data, e.g., the relations between pixels of an input image. It operates on a sequence of linear and nonlinear transformations  to infer a result, e.g., whether an input medical image indicates the patient has pneumonia. The linear transformations are in two typical forms: \emph{dot product} and \emph{convolution}. The nonlinear transformations leverage \emph{activations} such as the Rectified Linear Unit (\emph{ReLu}) to approximate complex functions~\cite{hornik1991approximation} and \emph{pooling} (e.g., max pooling and mean pooling) for dimensionality reduction. CNN repeats the linear and nonlinear transformations recursively to reduce the high-dimensional input data to a low-dimensional feature vector for classification at the \emph{fully connected layer}. Without losing generality, we use image classification as an example in the following discussion, aiming to provide a lucid understanding of the CNN architecture as illustrated in Fig.~\ref{overview}.

\emph{Convolutional Layer.} As shown in Fig.~\ref{overview:a}, the input to a convolutional layer has the dimensions of $w_i\times h_i\times c_i$, where $w_i$ and $h_i$ are the width and height of the input feature map and $c_i$ is the number of the feature maps (or channels). For the first layer, the feature maps are simply the input images. Hereafter, we use the subscript $i$ to denote input and $o$ output. The input is convolved with $c_o$ groups of kernels. The size of each group of kernel is $k_p\times k_q\times c_i$, in which $k_p$ and $k_q$ are the width and height of the kernel. The number of channels of the kernel group must match with the input, i.e., $c_i$. The convolution will produce the feature output, with a size of $w_o\times h_o\times c_o$. More specifically, the $(m,n)$-th element in the $t$-th $(1\leq{t}\leq{c_o})$ output feature is calculated as follows:
\begin{equation}
\small
z(m,n,t)=\sum_{j=1}^{c_i}\sum_{u=0}^{k_p-1}\sum_{v=0}^{k_q-1} k(u,v,j,t)x(m-u,n-v,j), \label{conv_eq}
\end{equation}
where $k$ and $x$ are the kernel and input, respectively. \emph{For the ease of description, we omit the bias in Eq. (\ref{conv_eq})}. Nevertheless, it can be easily transformed into the convolution or weight matrix multiplication \cite{han2016eie}.

The last convolutional layer is typically connected with the \emph{fully-connected layer}, which computes the weighted sum, i.e., a dot product between the weight matrix $w$ of size $n_o\times n_i$ and a flatten feature vector of size $n_i\times 1$. The output is a vector with the size of $n_o\times 1$. Each element of the output vector is calculated below:
\begin{equation}
\small
\vspace{-0.05in}
z(i)=\sum_{j=1}^{n_i} w(i,j)x(j). \label{weight_eq}
\end{equation}

\emph{Activation.} Nonlinear activation is applied to convolutional and weighted-sum outputs in an elementwise manner. In this work, we mainly target on the \textit{ReLu} activation function, $f(x)=\max\{0,x\}$, which is widely adopted in state-of-the-art neural networks such as AlexNet~\cite{krizhevsky2012imagenet} and VGG-16~\cite{simonyan2014very}.

\emph{Pooling.} Pooling conducts downsampling to reduce dimensionality.
In this work, we consider \emph{Mean pooling}, which is implemented in CryptoNets and also commonly adopted in state-of-art CNNs. It splits a feature map into regions and averages the regional elements. Compared to max pooling (another pooling function which selects the maximum value in each region), authors in~\cite{zhou2016learning} have claimed that while the max and mean pooling functions are rather similar,
the use of mean pooling encourages the network to identify the complete extent of the object,
which builds a generic localizable deep representation that exposes the implicit attention of CNNs on an image.


%
%
\subsection{Threat Model}
Similar to~\cite{liu2017oblivious,juvekar18gazelle,mohassel2017secureml,rouhani2018deepsecure}, we adopt the semi-honest model, in which both parties try to learn additional information from the message received (assuming they have bounded computational capability). That is, the client $\mathcal{C}$ and server $\mathcal{S}$ will follow the protocol, but $\mathcal{C}$ wants to learn the model parameters and $\mathcal{S}$ attempts to learn the data.
Hence, the goal is to make the server oblivious of the private data from the clients, and prevent the client from learning the model parameters of the server.  We would prove that the proposed framework is secure under semi-honest corruption using ideal/real security~\cite{goldreich2009foundations}.  Our framework targets to protect clients' sensative data, and service providers' models which have been trained by service providers with significant resources (e.g., private training data and computing power). Protecting models is usually sufficient through  protecting  the model parameters, which are the most critical information for a model. Moreover, many applications are even built on well-known  deep network structures such as AlexNet~\cite{krizhevsky2012imagenet}, VGG16/19~\cite{simonyan2014very} and ResNet50~\cite{he2016deep}.  Hence it is typically not necessary to protect the structure (number of layers, kernel size, etc). In the case that the implemented structure is proprietary and has to be protected,  service providers can introduce redundant layers and kernels to hide the real structure~\cite{liu2017oblivious, juvekar18gazelle}.


There is also an array of emerging attacks to the security and privacy of the neural networks~\cite{wang2018stealing,shokri2017membership,tramer2016stealing,fredrikson2015model,liu2017trojaning, hu2017generating}. They can be further classified by the processes that they are targeting at: training, inference (model) and input.

\emph{(1) Training.} The attack in~\cite{wang2018stealing} attempts to steal the hyperparameters during training. The membership inference attack~\cite{shokri2017membership} wants to find out whether an input belongs to the training set based on the similarities between models that are privately trained or duplicated by the attacker. This paper focuses on the inference stage and does not consider such attacks in training, since the necessary variables for launching these attacks have been released in memory and the training API is not provided.

\emph{(2) Model}. The model extraction attack~\cite{tramer2016stealing} exploits the linear transformation at the inference stage to extract the model parameters and the model inversion attack~\cite{fredrikson2015model} attempts to deduce the training sets by finding the input that maximizes the classification probability. The success of these attacks requires full knowledge of the softmax probability vectors. To mitigate them, the server can return only the predicted label but not the probability vector or limits the number of queries from the attacker. The Generative Adversarial Networks (GAN) based attacks \cite{liu2017trojaning} can recover the training data by accessing the model. In this research, since the model parameters are successfully protected from the clients, this attack can be defended effectively.

\emph{(3) Input.} A plethora of attacks adopt adversarial examples by adding a small perturbation to the input in order to cause the neural network to misclassify~\cite{hu2017generating}. Since rational clients pay for prediction services, it is not of their interest to obtain an erroneous output. Thus, this attack does not apply in our framework.

\subsection{Cryptographic Tools}
The proposed privacy-preserved deep neural network framework, i.e., CHEETAH, employs two fundamental cryptographic tools as outlined below.

\emph{(1) Packed Homomorphic Encryption}. 
%
Homomorphic Encryption (HE) is a cryptographic primitive that supports meaningful computations on encrypted data
without the decryption key. It has found increasing applications in data communication, storage and computation~\cite{takabi2010security}. Traditional HE operates on individual ciphertext~\cite{zhang2018gelu}, while the \emph{packed homomorphic encryption} (PHE) enables packing of multiple values into a single ciphertext and performs component-wise homomorphic computation in a Single Instruction Multiple Data (SIMD) manner~\cite{brakerski2013packed} to take the advantages of parallelism. Among various PHE techniques, our work builds on the private-key Brakerski-Fan-Vercauteren (BFV) scheme \cite{fan2012somewhat}, which involves four parameters\footnote{The readers are referred to \cite{juvekar18gazelle} for more detail.}: 1) ciphertext modulus $q$, 2) plaintext modulus $p$, 3) number of ciphertext slots $n$, and 4) a Gaussian noise with a standard deviation $\sigma$. The secure computation involves two parties, i.e., the client $\mathcal{C}$ and server $\mathcal{S}$.

In PHE, the encryption algorithm encrypts a plaintext message vector $x$ from $\mathbb{Z}^n$ into a ciphertext $[x]$ with $n$ slots. We denote $[x]_{\mathcal{C}}$ and $[x]_{\mathcal{S}}$ as the ciphertexts encrypted by client $\mathcal{C}$ and server $\mathcal{S}$, respectively. The decryption algorithm returns the plaintext vector $x$ from the ciphertext $[x]$. Computation can be performed on the ciphertext. In a general sense, an evaluation algorithm inputs several ciphertexts $[x_1],[x_2],\cdots$ and outputs a ciphertext $[x']=f([x_1],[x_2],\cdots)$. The function $f$ is constructed by homomorphic addition (Add), multiplication (Mult) and permutation (Perm). Add($[x]$,$[y]$) outputs a ciphertext $[x+y]$ which encrypts the elementwise sum of $x$ and $y$. Mult($[x]$,$u$) outputs a ciphertext $[x\circ{u}]$ which encrypts the elementwise multiplication of $x$ and plaintext $u$. It is worth pointing out that CHEETAH is designed to require multiplication between a ciphertext and a plaintext only, but not the much more expensive multiplication between two ciphertexts.
Perm($[x]$) permutes the $n$ elements in $[x]$ into another ciphertext $[x_\pi]$,  where $x_\pi=(x({\pi{_0}}),x({\pi{_1}}),\cdots)$ and $\pi_i$ is a permutation of  $\{0,1,\cdots,n-1\}$.

The run-time complexities of Add and Mult are significantly lower than Perm. From our experiments, one Perm is 56 times slower than one Add and 34 times slower than one Mult. This observation motivates the design of CHEETAH, which completely eliminates permutations in convolution and dot product transformations, thus substantially reducing the overall computation time.

It is worth pointing out that neural networks always deal with floating point numbers while the PHE is in the integer domain. Specifically, neural networks typically use real number arithmetic, not modular arithmetic. On the other hand, direct increasing plaintext modulus in PHE increases noise budget consumption, and also decreases the initial noise budget, which causes limited Homomorphic operations.
As for the original floating point numbers in neural networks, they are firstly quantized into 8-bit signed integers with fix-point encoding. As for the transformation from fix-point number to integer, our implementation adopts the highly efficient encoding for BFV in Microsoft SEAL library \cite{sealcrypto} to establish a mapping from real numbers in neural network to plaintext elements in PHE. This makes real number arithmetic workable in PHE without data overflow.
Thereafter, our design is described in floating point domain with real number input.

\emph{(2) Secret Sharing}. In the secret sharing protocol, a value is shared between two parties, such that combining the two secrets yields the true value \cite{riazi2018chameleon}. In order to additively share a secret $m$, a random number, $s$, is selected and two shares are created as $\langle{m}\rangle_0=s$ and $\langle{m}\rangle_1=m-s$. Here, $m$ can be either plaintext or ciphertext. A party that wants to share a secret sends one of the shares to the other party. To reconstruct the secret, one needs to only add two shares $m=\langle{m}\rangle_0+\langle{m}\rangle_1$.

While the overall idea of secret share (SS) is straightforward, creative designs are often required to enable its effective application in practice, because in many applications the two parties need to perform complex nonlinear computation on their respective shares and thus it is non-trivial to reconstruct the final result based on the computed shares. Due to this fundamental hurdle, the existing approaches discussed in Sec.~\ref{Intro:related} predominately chose to use GC, instead of SS, to implement the nonlinear functions.
However, GC is computationally costly for large input~\cite{rouhani2018deepsecure,woodruff2007revisiting,juvekar18gazelle}. 
Specifically, our benchmark shows that GC takes about 263 seconds to compute a nonlinear ReLu function with 3.2M input values, which is part of the VGG-16 framework \cite{simonyan2014very}.
In this work, we propose a creative PHE-based SS for CHEETAH to implement secret nonlinear computation, which requires only $\frac{1}{2}$ round communication for each nonlinear function
, thus achieving multiple orders of magnitude reduction of the computation time. For example, CHEETAH achieves a speedup of 1793 times over GAZELLE in computing the nonlinear ReLu function. 

\section{Design of Privacy Preserved Inference}\label{systm}
A neural network is organized into layers. For example, CNN consists of convolutional layers and fully-connected dense layers. Each layer includes linear transformation (i.e., weighted sum for a fully-connected dense layer or convolution for a convolutional layer), followed by nonlinear tranformation (such as activation and pooling). All existing schemes intend to securely compute the results for linear transformation first, and then perform the nonlinear computation. Although it appears logical, such design leads to a fundamental performance bottleneck as discussed in Sec.~\ref{Intro}. The proposed approach, CHEETAH, is based on a creative design, named {\em joint obscure neural computing}, which only computes a \emph{partial} linear transformation output and uses it to complete the nonlinear transformation. It achieves several orders of magnitude speedup compared with existing schemes.

We introduce the basic idea of CHEETAH via a simple example based on a two-layer CNN (with a convolutional layer and a dense layer), which can be formulated as follows:
\begin{equation}
z=w\cdot{f}(k\ast{x}),\label{eq:1}
\end{equation}
where $f(\cdot)$ is the activation function, $x$ is the $2\times 2$ input data, $k$ is a $3\times 3$ kernel for the convolutional layer, $\ast$ stands for convolution and $w$ is the weight matrix for the dense layer:
\mathleft
\begin{equation*}
{\footnotesize
x= \left[\begin{array}{cc}
x(1,1)&x(1,2)\\
x(2,1)&x(2,2)
\end{array} \right],
k=\left[\begin{array}{ccc}
k(1,1)&k(1,2)&k(1,3)\\
k(2,1)&k(2,2)&k(2,3)\\
k(3,1)&k(3,2)&k(3,3)
\end{array} \right]}\,\textrm{and}
\end{equation*}
\begin{equation*}
{\footnotesize
w=\left[\begin{array}{cccc}
w(1,1)&w(1,2)&w(1,3)&w(1,4)\\
w(2,1)&w(2,2)&w(2,3)&w(2,4)
\end{array} \right]}.
\end{equation*}
\mathcenter
Note that while we use the simple two-layer CNN to lucidly describe the main idea, CHEETAH is actually applicable to any neural networks with any layer structure and input data size.  In the rest of this section, we first present CHEETAH for a Single Input Single Output (SISO) convolution layer and then discuss the cases for Multiple Input Multiple Output (MIMO) convolution and fully connected dense layers.
\subsection{SISO Convolutional Layer}\label{sec.3.1}
The process of convolution can be visualized as placing the kernels at different locations of the input data. At each location, an element-wise sum of product is computed between the kernel and corresponding data values. If the convolution of the above example, i.e., $k\ast{x}$, is computed in plaintext, the result, denoted as $Con$, should include four elements, $Con=[Con_1,Con_2,Con_3,Con_4]$:
\begin{equation*}
{\footnotesize
\begin{array}{c}
Con_1: k(2,2)x(1,1)+k(2,3)x(1,2)+k(3,2)x(2,1)+k(3,3)x(2,2),\\
Con_2: k(2,1)x(1,1)+k(2,2)x(1,2)+k(3,1)x(2,1)+k(3,2)x(2,2),\\
Con_3: k(1,2)x(1,1)+k(1,3)x(1,2)+k(2,2)x(2,1)+k(2,3)x(2,2),\\
Con_4: k(1,1)x(1,1)+k(1,2)x(1,2)+k(2,1)x(2,1)+k(2,2)x(2,2).
\end{array}}
\end{equation*}

In the problem setting of secure MLaaS (as introduced in Sec.~\ref{prery}), the client $\mathcal{C}$ owns the data $x$, while the server $\mathcal{S}$ owns the CNN model (including $k$ and $w$). The goal is to ensure that the server does not have access to $x$ and the client cannot learn the server's model parameters. To this end, in GAZELLE, $\mathcal{C}$ encrypts $x$ into $[x]_{\mathcal{C}}$ by using HE and sends it to $\mathcal{S}$.
In the following discussion, both server and client use private-key BFV encryption \cite{fan2012somewhat}.
The subscript $[\cdot ]_{\mathcal{C}}$ denotes ciphertext encrypted by the client's private key, while $[\cdot ]_{\mathcal{S}}$ denotes ciphertext encrypted by the private key of server.

$\mathcal{S}$ performs HE computation to calculate the convolution $k\ast{[x]_{\mathcal{C}}}$. To accelerate the computation, packed HE is employed. For example, to compute the first element of the convolution (i.e., $Con_1$), a single cipheretxt can be created to contain the vector $[x(1,1), x(1,2), x(2,1), x(2,2)]_{\mathcal{C}}$. On the other hand, a packed plaintext vector is created for $[k(2,2),k(2,3),k(3,2),k(3,3)]$. The packed HE supports the computation of element-wise multiplication between the two vectors in a single operation, yielding a single ciphertext for the vector $[k(2,2)x(1,1),k(2,3)x(1,2),\cdots,k(3,3)x(2,2)]_{\mathcal{C}}$. However, we still need to add the vector's elements together to compute $Con_1$. Since the vector is in a single ciphertext, direct addition is not possible.  GAZELLE uses permutation (Perm) to compute the sum~\cite{juvekar18gazelle}.
{\color{black}For example, given a ciphertext that has four elements, it is firstly permed such that the last two elements are moved to the first two slots in the ciphertext. Then the permed ciphertext is added with the original counterpart, which results in a ciphertext whose first two elements are the partial sum of the four elements. Then that added ciphertext is permed such that its second element is moved to the first slot. The sum of the four elements is obtained by adding the permed ciphertext and the non-permed one. The resultant sum is at the first slot of the final ciphertext.}

However, computing the sum using Perm is costly, with the complexity of $O(r^2)$ for convolution and $O(\log{\frac{n}{n_o}}+\frac{n_in_o}{n})$ for weighted sum in the dense layer, where $n_o$, $n_i$ and $r$ are the output dimension, input dimension, and kernel size, respectively).
From our experiments, one Perm is 56 times slower than one Add and 34 times slower than one Mult.

\begin{figure}[!btp]
\centering
\includegraphics[scale=0.7]{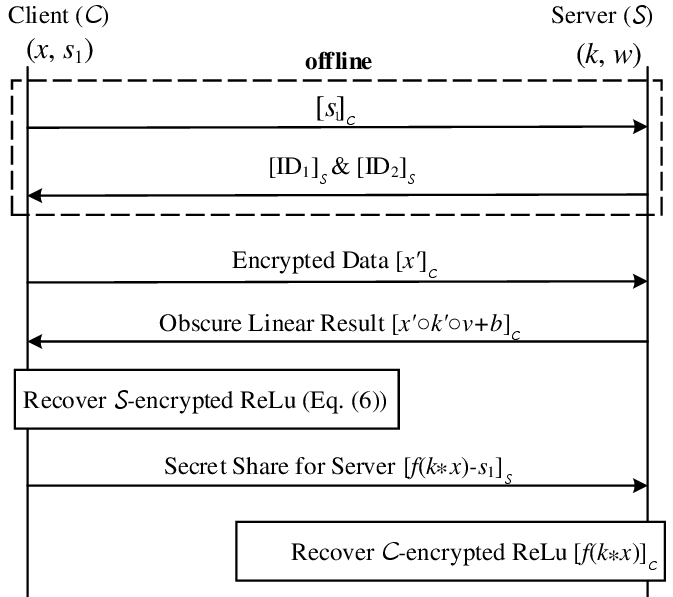}
\caption{The overall design of CHEETAH.}\vspace*{-0.2in}
\label{syst}
\end{figure}

In this paper, we propose a novel idea to enable an incomplete (or obscure) linear transformation result to propagate to the next nonlinear transformation to continue the neural computation, thus eliminating the need for ciphertext permutations. The overall design is motivated by the double-secret scheme for solving linear system of equations  \cite{du2001privacy}. Our scheme is illustrated in Fig.~\ref{syst}.

\vspace*{0.05in}\noindent{\bf (1) Packed HE Encryption.} $\mathcal{C}$ and $\mathcal{S}$  transform the data $x$ and kernel $k$ into $x'$ and $k'$, respectively, as follows:
{\small
\begin{equation*}
\begin{split}
x'=[x(1,1),x(1,2),x(2,1),x(2,2),x(1,1),x(1,2),x(2,1),x(2,2),&
\\
x(1,1),x(1,2),x(2,1),x(2,2),x(1,1),x(1,2),x(2,1),x(2,2)],&
\end{split}
\end{equation*}}
{\small
\begin{equation*}
\begin{split}
k'=[k(2,2),k(2,3),k(3,2),k(3,3),k(2,1),k(2,2),k(3,1),k(3,2),&
\\
k(1,2),k(1,3),k(2,2),k(2,3),k(1,1),k(1,2),k(2,1),k(2,2)].&
\end{split}
\end{equation*}}

\begin{figure}[b]
\vspace*{-0.17in}
\centering
\includegraphics[width=1\columnwidth]{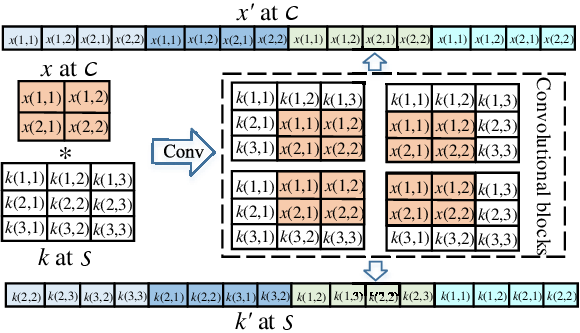}
\caption{Data transformation at client and server.}
\label{conv_trans}
\end{figure}
As illustrated in Fig.~\ref{conv_trans}, four convolutional blocks are computed. For example, the first convolutional block computes {\footnotesize$x(1,1)\times{k(2,2)}+x(1,2)\times{k(2,3)}+x(2,1)\times{k(3,2)}+x(2,2)\times{k(3,3)}$}. The elements in each convolutional block are sequentially extracted into a packed ciphertext $[x']_{\mathcal{C}}$. Meanwhile, $\mathcal{S}$ also transforms the kernel $k$ into $k'$ according to each convolutional block. Note that the transformation is completed offline. $\mathcal{C}$ encrypts $x'$ and sends $[x']_{\mathcal{C}}$ to $\mathcal{S}$.


\vspace*{0.03in}\noindent{\bf (2) Perm-free Secure Linear Computation}. Upon receiving $[x']_{\mathcal{C}}$, $\mathcal{S}$ performs the linear computation based on the client-encrypted data. A distinguished feature of the proposed design is to eliminate the costly permutations.

Let $x'\circ{k'}$ denote the elementwise multiplication between $x'$ and $k'$. As we can see, the sum of four elements for each block in $x'\circ{k'}$ corresponds to one element of the convolution result. For example, the four elements for first block, i.e., {\footnotesize$[x(1,1),x(1,2),x(2,1),x(2,2)]$} and {\footnotesize$[k(2,2), k(2,3),k(3,2),$ $k(3,3)]$}, correspond to $Con_1$. The next block (i.e., {\footnotesize$[x(1,1),x(1,2),$$x(2,1),$ $x(2,2)]$} and {\footnotesize$[k(2,1),k(2,2),k(3,1),k(3,2)]$}) correspond to $Con_2$, and so on and so forth.

$\mathcal{S}$ performs Mult$([x']_{\mathcal{C}},k')$ to obtain $[x'\circ{k'}]_{\mathcal{C}}$. The result is the client-encrypted elementwise multiplication between $x'$ and $k'$. But $\mathcal{S}$ does not intend to calculate the sum of each block to obtain the final convolution result as GAZELLE does, because it would need the costly permutations. Instead, it intends to let $\mathcal{C}$ decrypt $[x'\circ{k'}]_{\mathcal{C}}$ to compute the sum in the plaintext.

However,  naively sending $[x'\circ{k'}]_{\mathcal{C}}$ to the client would allow the client to obtain the neural network model information, i.e., $k$. To this end, $\mathcal{S}$ disturbs each element of the convolution result with a randomly multiplicative blinding factor. Specifically, $\mathcal{S}$ pre-generates
a pair of random numbers that satisfy $v_{i1}v_{i2}=1$, for each $i$-th to-be-summed block in $[x'\circ{k'}]_{\mathcal{C}}$, where $i\in\{1,2,3,4\}$ in this example.

$\mathcal{S}$ constructs the following vector $v$ by using $v_{i1}$:
\begin{equation*}
\begin{split}
v=[v_{11},v_{11},v_{11},v_{11},v_{21},v_{21},v_{21},v_{21},&\\
v_{31},v_{31},v_{31},v_{31},v_{41},v_{41},v_{41},v_{41}],&
\end{split}
\end{equation*}
which will be used to scramble $[x'\circ{k'}]_{\mathcal{C}}$ by multiplying it with $v$ before it is sent to $\mathcal{C}$. Note that, as each individual in the $i$-th four-element block is multiplied with the same factor (since $v_{11},v_{21},v_{31},v_{41}$ are repeated four times in $v$), it would leak the relative magnitude among those four elements in each block.
{\color{black}To this end, $\mathcal{S}$ further constructs a noise vector as follows:
\begin{equation*}
\begin{split}
b=[b_{11},b_{12},b_{13},b_{14},
b_{21},b_{22},b_{23},b_{24},&\\
b_{31},b_{32},b_{33},b_{34},
b_{41},b_{42},b_{43},b_{44}]&,
\end{split}
\end{equation*}
where $b_{ij}$ are random numbers subject to $\sum_{j=1}^{4}b_{ij}=v_{i1}\delta_i$ that is uniformly distributed in $[-\epsilon,\epsilon]$ where $\epsilon$ is a model parameter known to server $\mathcal{S}$.}

At the same time, $\mathcal{S}$ uses $v_{i2}$ to create the following vectors:
\begin{equation*}
\begin{array}{c}
{\rm ID}_{1}=[{\rm ID}_{11},{\rm ID}_{21},{\rm ID}_{31},{\rm ID}_{41}],\\
{\rm ID}_{2}=[{\rm ID}_{12},{\rm ID}_{22},{\rm ID}_{32},{\rm ID}_{42}],\label{eq:ID2}\\
\end{array}
\end{equation*}
where $({\rm ID}_{i1},{\rm ID}_{i2})$ is a pair of \emph{polar indicator},
\begin{equation}\label{indca}
({\rm ID}_{i1},{\rm ID}_{i2}) = \left\{ {\begin{array}{*{20}{c}}
{(0,v_{i2}),{\rm\;if\;}v_{i1} > 0}\\
{(v_{i2},-v_{i2}),{\rm\;if\;}v_{i1} <0}.
\end{array}} \right.
\end{equation}
$\mathcal{S}$ encrypts ${\rm ID}_{1}$ and ${\rm ID}_{2}$ by using packed HE. The encrypted values, i.e., $[{\rm ID}_1]_{\mathcal{S}}$ and $[{\rm ID}_2]_{\mathcal{S}}$, will be sent to $\mathcal{C}$ for the nonlinear computation as to be discussed later.
Note that, $[{\rm ID}_1]_{\mathcal{S}}$ and $[{\rm ID}_2]_{\mathcal{S}}$ can be transmitted to $\mathcal{C}$ offline, as $v_{i1}$ and $v_{i2}$ are pre-generated by $\mathcal{S}$.

Now, let us put all pieces together for the secure computation of convolution: $\mathcal{C}$ encrypts $x'$ and sends $[x']_{\mathcal{C}}$ to $\mathcal{S}$. $\mathcal{S}$ pre-computes $v\circ{k'}$ in plaintext and then multiplies the result with $[x']_{\mathcal{C}}$ to obtain $[x'\circ{k'}\circ{v}]_{\mathcal{C}}$. As we can see, the $i$-th convolution element (which corresponds to the sum of $i$-th four-element block in $[x'\circ{k'}\circ{v}]_{\mathcal{C}}$) is actually multiplied with a random number $v_{i1}$.
{\color{black}Finally, $\mathcal{S}$ adds the noise vector by Add$([x'\circ{k'}\circ{v}]_{\mathcal{C}},b)=[x'\circ{k'}\circ{v}+b]_{\mathcal{C}}$. In this way, $b$ disturbs each element of convolution result (the sum of four elements in each block) with a random noise $\delta_i$ while $v$ scales each noised element.}

Next, we will show that, although the convolution result is not explicitly calculated, the partial (obscure) result, i.e., $[x'\circ{k'}\circ{v}+b]_{\mathcal{C}}$, is sufficient to compute the nonlinear transformation (e.g., activation and pooling).

\vspace*{0.03in}\noindent{\bf (3) PHE-based Secret Share for Non-Linear Transformation}. $\mathcal{S}$ sends $[x'\circ{k'}\circ{v}+b]_{\mathcal{C}}$, $[{\rm ID}_1]_{\mathcal{S}}$ and $[{\rm ID}_2]_{\mathcal{S}}$ to $\mathcal{C}$ (note that $[{\rm ID}_1]_{\mathcal{S}}$ and $[{\rm ID}_2]_{\mathcal{S}}$ are transmitted to $\mathcal{C}$ offline).

$\mathcal{C}$ decrypts $[x'\circ{k'}\circ{v}+b]_{\mathcal{C}}$ and sums up each four-element block in plaintext, yielding $y=[y{(1)},y{(2)},y{(3)},y{(4)}]$. It is not difficult to show that $y(i)$ is $v_{i1}$ times of the disturbed convolution,
i.e., $y(i)= v_{i1}\times(Con_i+\delta_i)$.

If $\mathcal{C}$ had the true convolution outcome, i.e., $Con_i$, it would compute the ReLu function as follows:
\begin{equation}
f_R(Con_i) = \left\{ {\begin{array}{*{20}{c}}
{Con_i,{\rm\;if\;}Con_i \geq 0}\\
{0,{\rm\;if\;}Con_i < 0}.
\end{array}} \right.
\end{equation}

However, $\mathcal{C}$ only has $y(i)= v_{i1}\times(Con_i+\delta_i)$. Since $v_{i1}$ is a random number that could be positive or negative, it is infeasible to obtain correct activation directly. Instead, $\mathcal{C}$ computes
\begin{equation}
{\rm Add}({\rm Mult}([{\rm ID}_1]_{\mathcal{S}},y),{\rm Mult}([{\rm ID}_2]_{\mathcal{S}},f_R(y))).\label{eq5}
\end{equation}

We can show that the above calculation essentially recovers the server-encrypted ReLu function of $Con_i+\delta_i$, i.e., $[f(k\ast{x}+\delta)]_{\mathcal{S}}$ where $\delta=\{\delta_i\}$. Since $y(i)= v_{i1}\times(Con_i+\delta_i)$, $f_R(y(i))$ may yield four possible outputs, depending on the signs of $v_{i1}$ and $Con_i+\delta_i$.
\begin{equation}
f_R(y(i)) = \left\{ {\begin{array}{*{20}{l}}
{y(i),{\rm\;if\;}v_{i1} > 0\;\&\;(Con_i+\delta_i) \geq 0}\\
{y(i),{\rm\;if\;}v_{i1} <0\;\&\;(Con_i+\delta_i) < 0}\\
{0,{\rm\;if\;}v_{i1} > 0\;\&\;(Con_i+\delta_i) < 0}\\
{0,{\rm\;if\;}v_{i1} <0\;\&\;(Con_i+\delta_i) \geq 0}.
\end{array}} \label{eq:6}\right.
\end{equation}
{\color{black}For example, when $v_{i1} > 0$ and $(Con_i+\delta_i)\geq 0$, we have ${\rm ID}_1=\{0\}$ according to Eq.~(4)
and thus ${\rm Mult}([{\rm ID}_1]_{\mathcal{S}},y)=[0]_{\mathcal{S}}$. On the other hand, ${\rm ID}_2=[v_{12},v_{22},v_{32},v_{42}]$. Since $y(i)= v_{i1}\times(Con_i+\delta_i)$, we have ${\rm Mult}([{\rm ID}_2]_{\mathcal{S}},f_R(y))=[v_{12}v_{11}(Con_1+\delta_1),\cdots,v_{42}v_{41}(Con_4+\delta_4)]_{\mathcal{S}}$. Note that we have chosen $v_{i1}v_{i2} = 1$. Therefore, Eq.~(\ref{eq5}) should yield $[Con_1+\delta_1,Con_2+\delta_2,Con_3+\delta_3,Con_4+\delta_4]_{\mathcal{S}}$. This is clearly the server-encrypted ReLu output of $Con+\delta$. Similarly, we can examine other cases of $v_{i1}$ and $Con_i+\delta_i$ in Eq.~(\ref{eq:6}) and show that Eq.~(\ref{eq5}) always produce the server-encrypted ReLu outcome $f(k\ast{x}+\delta)$. We will show in Sec.~\ref{perf} that the ReLu function of noised linear result introduces negligible accuracy loss to the neural networks while $\delta_i$ and $v_{1i}$ prevent client from inferring the right $Con_i$.}


Subsequently, $\mathcal{C}$ creates a ReLu share $s_1$ and computes the server's share as Add$([f(k\ast{x}+\delta)]_{\mathcal{S}},-s_1)=[f(k\ast{x}+\delta)-s_1]_{\mathcal{S}}$. $\mathcal{C}$ sends it along with $[s_1]_{\mathcal{C}}$ (i.e., the client-encrypted share $s_1$, which can be pre-generated by $\mathcal{C}$) to $\mathcal{S}$.

$\mathcal{S}$ decrypts $[f(k\ast{x}+\delta)-s_1]_{\mathcal{S}}$ to obtain a share of the plaintext activation result, i.e., $f(k\ast{x}+\delta)-s_1$.  It then computes Add$([s_1]_{\mathcal{C}},f(k\ast{x}+\delta)-s_1)$ to obtain $[a]_{\mathcal{C}}=[f(k\ast{x}+\delta)]_{\mathcal{C}}$, i.e., the client-encrypted nonlinear transformation result. Note that, the introduce of $\delta$ dose not effect the neural network performance as shown in Sec.~\ref{perf}.

Till now, the computation of the current layer (including linear convolution and nonlinear activation) is completed. The output of this layer (i.e., $[f(k\ast{x}+\delta)]_{\mathcal{C}}$) will serve as the input for the next layer. If the next layer is still convolution, the server simply repeats the above process. Otherwise, if the next is a fully-connected dense layer, a similar approach can be taken as to be discussed in Sec.~\ref{sec:3.3}.

Note that some CNN models employ pooling after activation to reduce its dimensionality. For example, mean pooling takes the activations as the input, which is divided into a number of regions. The averaged value of each region is used to represent that region. Both $\mathcal{C}$ and $\mathcal{S}$ can respectively average their activation shares (i.e., $s_1$ and $f(k\ast{x}+\delta)-s_1$) to obtain the share of mean pooling. Meanwhile, a similar scheme can be applied if the bias is included.

%

\subsection{MIMO Convolutional Layer}
The above SISO method can be readily extended to MIMO convolutional layer in order to process multiple inputs simultaneously. Assume there are $c_i$ input data (i.e., $x$). Let $c_n$ be the number of input data that can be packed into one ciphertext. Recall that each $x$ must be transformed to $x'$ as discussed in Sec.~\ref{sec.3.1}. Let $c_o$ denote the number of kernels and $r$ the size of each kernel. After transformation, the size of $x'$ is $r^2$ times of the original $x$. Therefore, each ciphertext can hold $c_n/r^2$ such transformed input data. Accordingly, the $c_i$ input data are transformed and encrypted into ${c_ir^2}/{c_n}$ ciphertexts.

The remaining process for linear and nonlinear computation is similar to SISO, except that the computation on a ciphertext actually calculates multiple input data simultaneously and that the convolution of all input ciphertexts based on one kernel are combined into one output ciphertext, yielding a total of $c_o$ output ciphertexts. MIMO is obviously more efficient in processing batches of input data.
\subsection{Fully-connected Dense Layer}\label{sec:3.3}
In a fully-connected dense layer, $\mathcal{S}$ uses the output of the previous layer (i.e., $[a]_{\mathcal{C}}$) to compute the weighted sum. Take the simple two-layer CNN as an example, the weighted sum computes
{\footnotesize\begin{equation*}
\begin{array}{c}
c_1=w(1,1)[a(1)]_{\mathcal{C}}+w(1,2)[a(2)]_{\mathcal{C}}+w(1,3)[a(3)]_{\mathcal{C}}+w(1,4)[a(4)]_{\mathcal{C}},\\
c_2=w(2,1)[a(1)]_{\mathcal{C}}+w(2,2)[a(2)]_{\mathcal{C}}+w(2,3)[a(3)]_{\mathcal{C}}+w(2,4)[a(4)]_{\mathcal{C}}.
\end{array}
\end{equation*}}

\noindent{The computation of $c_1$ and $c_2$ is intrinsically the same as the computation of each convolution element (i.e., $Con_1, \dots,$ $Con_4$) as discussed above.}

\subsection{Complexity Analysis}
In this subsection, we analyze the computation and communication cost of CHEETAH and compare it with other schemes.

\noindent{\bf (1) Computation Complexity}. The analysis of the computation complexity focuses on the number of ciphertext permutations (Perm), multiplications (Mult), and additions (Add). The notations to be used in the analysis are summarized as follows:
\begin{itemize}
\item $n$ is the number of slots in a ciphertext.
\item $q$ is the ciphertext space.
\item $n \log q$ is the number of bits of a ciphertext.
\item $n_i$ is the input dimension of a fully connected layer.
\item $n_o$ is the output dimension of a fully connected layer.
\item $r$ is the kernel size.
\item $c_i$ is the number of input data (channels) in MIMO.
\item $c_o$ is the number of kernels or the number of output feature maps in MIMO.
\item $c_n$ is the number of input data that can be packed into one ciphertext.
\end{itemize}

In SISO, recall that a ciphertext $[x']_{\mathcal{C}}$ is firstly sent to $\mathcal{S}$. $\mathcal{S}$ conducts one ciphertext multiplication and addition to get $[v\circ{k'}\circ x'+b]_{\mathcal{C}}$. Then $\mathcal{C}$ receives $[v\circ{k'}\circ x'+b]_{\mathcal{C}}$, performs the decryption, and gets the summed convolution $y$ in plaintext,
which is followed by 2 multiplications and 1 addition to get the encrypted ReLu, according to Eq. \ref{eq5}. Finally, $\mathcal{C}$ does another addition namely Add$([f(k\ast{x}+\delta)]_{\mathcal{S}},-s_1)$ to generate $\mathcal{S}$'s ReLu share. $\mathcal{S}$ finaly recovers the encrypted nonlinear result with another addition. Therefore,  total  3 multiplications and 4 additions are required in SISO. The complexity is $O(1)$.

In MIMO, $\mathcal{C}$ sends $\mathcal{S}$ ${c_ir^2}/{c_n}$ ciphertexts. Then $\mathcal{S}$ performs ${c_ir^2}/{c_n}$ Mult and $({c_ir^2}/{c_n}-1)$ Add to get an incomplete ciphertext for each of $c_o$ kernels. After that, each of $c_o$ incomplete ciphertext is added with noise vector by one addition. Then $\mathcal{S}$ sends those $c_o$ cipheretxts to $\mathcal{C}$, which decrypts them and obtain $c_o$ output features, creating $c_o/c_n$ plaintext. Based on Eq. \eqref{eq5}, $\mathcal{C}$ gets the encrypted ReLu with ${2c_o}/{c_n}$ multiplications and ${c_o}/{c_n}$ additions, because each of $c_o/c_n$ plaintext associates with $2$ multiplications and 1 addition. Finally, $\mathcal{C}$ performs another addition on each of ${c_o}/{c_n}$ ReLu ciphertexts to generate the ReLu share for $\mathcal{S}$. $\mathcal{S}$ then gets its ReLu share by decryption and recovers the nonlinear result by ${c_or^2}/{c_n}$ Add. Therefore, MIMO needs $(\frac{c_ic_or^2}{c_n}+\frac{2c_o}{c_n})$ multiplications and $(\frac{(c_i+1)c_or^2}{c_n}+\frac{2c_o}{c_n})$ additions, both with the complexity of $O(\frac{c_ic_or^2}{c_n})$.


In a fully-connected (FC) dense layer, $\mathcal{S}$ conducts ${n_in_o}/{n}$ multiplications to get ${n_in_o}/{n}$ intermediate ciphertext, where $n$ is usually much larger than $n_i$ and $n_o$. After that, the zero-sum vector is added on each of ${n_in_o}/{n}$ intermediate ciphertext to form $[x'\circ{w'}\circ{v}+b']_{\mathcal{C}}$\footnote{The structure of $b'$ is similar with $b$.} which is sent to $\mathcal{C}$. $\mathcal{C}$ does the decryption and gets the summed result in plaintext. Then $\mathcal{C}$ calculates the encrypted ReLu with 2 multiplications and 1 addition by Eq. \eqref{eq5}. Finally, one addition is performed to generate the ReLu share for $\mathcal{S}$, and $\mathcal{S}$ needs another Add to recover the encrypted nonlinear result. So the FC layer needs $(\frac{n_in_o}{n}+2)$ multiplications and $(\frac{n_in_o}{n}+3)$ additions, resulting in the complexity of $O(\frac{n_in_o}{n})$.

Table~\ref{sisocp} compares the computation complexity between CHEETAH and other schemes. Specifically, In the SISO case, CHEETAH (CH) has a constant complexity without permutation while GAZELLE (GA) has the complexity  $r^2$. In the MIMO case, GAZELLE has two traditional options for permutation, i.e., Input Rotation (IR) and Output Rotation (OR) \cite{juvekar18gazelle}. CHEETAH eliminates the expensive permutation without incurring more multiplications and additions, thus yielding a considerable gain. In the FC layer, we compare CHEETAH with a naive method (NA) in \cite{juvekar18gazelle} (the baseline of GAZELLE), Halevi-Shoup (HS) \cite{halevi2014algorithms} and GAZELLE. Through the obscure matrix calculation, obscure HE and secret share, CHEETAH further reduces the complexity of addition by $O(\log{\frac{n}{n_o}})$ compared to GAZELLE. In particular, $n$ is usually much larger than $n_o$, which makes this reduction significant. It is worth pointing out that CHEETAH completes both the linear and nonlinear operations with the above complexity while  the existing schemes such as GAZELLE  only finish the linear operation.
\begin{table}[!t]
\renewcommand\arraystretch{1.2}
\centering
{\scriptsize
\caption{Comparison of computation complexity.}
\vspace*{-0.13in}
\begin{tabular}{cccc}
  \hline
  \hline
 Method & Perm & Mult & Add\\
  \hline \hline
  GA-SISO& $O(r^2)$ & $O(r^2)$ & $O(r^2)$\\
  \textbf{CH-SISO} & $\bm0$ & $\bm{O(1)}$ & $\bm{O(1)}$\\
  \hline
  IR-MIMO & $O(c_ir^2)$ & $O(\frac{c_ic_or^2}{c_n})$ & $O(\frac{c_ic_or^2}{c_n})$\\
  OR-MIMO & $O(\frac{c_ic_or^2}{c_n})$ & $O(\frac{c_ic_or^2}{c_n})$ & $O(\frac{c_ic_or^2}{c_n})$\\
  \textbf{CH-MIMO} & $\bm0$ & $\bm{O(\frac{c_ic_or^2}{c_n})}$ & $\bm{O(\frac{c_ic_or^2}{c_n})}$\\
   \hline
  NA-FC\cite{juvekar18gazelle} & $O(n_o\log{n_i})$ & $O(n_o)$ & $O(n_o\log{n_i})$\\
  HS-FC\cite{halevi2014algorithms} & $O(n_i)$ & $O(n_i)$ & $O(n_i)$\\
  GA-FC & $O(\log{\frac{n}{n_o}}+\frac{n_in_o}{n})$ & $O(\frac{n_in_o}{n})$ & $O(\log{\frac{n}{n_o}}+\frac{n_in_o}{n})$\\
  \textbf{CH-FC} & \textbf{0} & $\bm{O(\frac{n_in_o}{n})}$ & $\bm{O(\frac{n_in_o}{n})}$\\
  \hline
  \hline
\end{tabular}
\label{sisocp}
\vspace*{-0.1in}
}
\end{table}

\noindent{\bf (2) Communication Complexity}. In the SISO case, CHEETAH has two transmissions: 1) $\mathcal{C}$ sends the encrypted data $[x']_{\mathcal{C}}$ to $\mathcal{S}$; 2) $\mathcal{S}$ sends $[x'\circ{k'}\circ{v}+b]_{\mathcal{C}}$ to $\mathcal{C}$. Thus the communication cost is $2n\log{q}$ bits. Note that the third transmission in Fig.~\ref{conv_trans} where $\mathcal{C}$ sends the encrypted ReLu share to $\mathcal{S}$ is the beginning of the next layer.

Similarly, in MIMO, the two transmissions are 1) $\mathcal{C}$ sends $\mathcal{S}$ ${c_ir^2}/{c_n}$ ciphertexts for $c_i$ input images; 2) $\mathcal{S}$ sends $\mathcal{C}$ $c_o$ cipheretexts for $c_o$ kernels. Note that, in the first transmission, ${c_ir^2}/{c_n}$ ciphertexts are transmitted at the first convolutional layer while only ${c_i}/{c_n}$ ciphertexts are needed in other layers. This is because the size of $\mathcal{S}$-encrypted ReLu will not be changed. In the second transmission, since $\mathcal{S}$ can simultaneously send each of $c_o$ cipheretexts after each calculation, the actual communication cost is on transmitting the last one of $c_o$ ciphertexts. Thus, CHEETAH has a pipelined communication cost as $(\frac{c_i}{c_n}+1)n\log{q}$ bits.

In the FC layer, the two transmissions are 1) $\mathcal{C}$ sends $\mathcal{S}$ an input ciphertext; 2) $\mathcal{S}$ sends $\mathcal{C}$ $n_in_o/n$ cipheretexts. As each of $n_in_o/n$ cipheretexts can be simultaneously transmitted after each calculation, the actual communication cost is the last one of $n_in_o/n$ ciphertexts. The total pipelined cost is thus $2n\log{q}$ bits. The quantitative communication comparison to other approaches is given in Sec.~\ref{perf}.
\section{Security Analysis}\label{secr}
{\color{black}We follow the ideal/real world paradigm~\cite{goldreich2009foundations,mishra2020delphi,zheng2020securely} to prove the security of CHEETAH. We start with defining the ideal functionality $f^{\rm OMI}$ which captures the security properties we want to achieve for Outsourced MLaaS Inference.

\vspace*{0.05in}
\noindent\textbf{Defintion 1.}
\textit{The ideal functionality $f^{\rm OMI}$ of outsourced MLaaS inference consists of the following parts:}

\textit{- \textbf{Input.} The server sends model parameters $\bm{M}$, e.g., kernel $k\in\bm{M}$, to $f^{\rm OMI}$. The
client sends private input $x$ to $f^{\rm OMI}$.}

\textit{- \textbf{Computation.} Upon receiving the model parameters from server and the private input $x$ from client, $f^{\rm OMI}$ conducts MLaaS inference by linear and nonlinear computation with $x$ and produces the nonlinear result $f(x\ast{k})=ReLu(x\ast{k})$.}

\textit{- \textbf{Output:} The $f^{\rm OMI}$ sends respective share of the nonlinear result $f(x\ast{k})=ReLu(x\ast{k})$ to client and server. As for the last layer, the $f^{\rm OMI}$ sends the obscure linear result to client with one random number in $v$.}

\vspace*{0.08in}
Given the ideal functionality $f^{\rm OMI}$, we give the formal security definition as follows.

\vspace*{0.08in}
\noindent{\textbf{Definition 2.}} \textit{A protocol $\Pi$ securely computes the $f^{\rm OMI}$ in the semi-honest adversary setting with static corruption if it provides the following guarantees:}

\textit{- \textbf{Corrupted server.} We require that a corrupted and semi-honest
server does not learn any information about the
values in the client's private input $x$. Formally, there should exist a Probabilistic
Polynomial Time (PPT) simulator $sim_{\mathcal{S}}$ such that $view_{\mathcal{S}}^\Pi\overset{c}{\approx}{sim_{\mathcal{S}}(\bm{M},out)}$,
where $view_{\mathcal{S}}^\Pi$ denotes the view of the server in the real protocol execution (including the server's input, randomness, and the transcript of the protocol). $sim_{\mathcal{S}}(\bm{M},out)$ is the simulation based on $\mathcal{S}$'s input, i.e., $\bm{M}$, and its final output `$out$', e.g., the share of nonlinear function. The ``$\overset{c}{\approx}$'' denotes ``computationally indistinguishable''.}

\textit{- \textbf{Corrupted client.} We require that a corrupted and semi-honest client does not learn any information about the
server's model parameters beyond some generic meta-parameters, i.e, the number of input and output channels and the number of layers. Formally, there should exist a PPT simulator $sim_{\mathcal{C}}$ such that $view_{\mathcal{C}}^\Pi\overset{c}{\approx}{sim_{\mathcal{C}}(x,out)}$,
where $view_{\mathcal{C}}^\Pi$ denotes the view of the client in the real protocol execution (including the client's input, randomness, and the transcript of the protocol). $sim_{\mathcal{C}}(x,out)$ is the simulation based on $\mathcal{C}$'s input, i.e., $x$, and its final output `$out$', e.g., the share of nonlinear function.}

\vspace*{0.08in}
\noindent\textbf{Theorem 1.} \textit{Our protocol provides a secure realization of the ideal functionality $f^{\rm{ODT}}$ according to \rm{Definition 2}.}

\vspace*{0.08in}
\noindent\textit{Proof.} According to our security definition, we need to show a simulator for different corrupted parties i.e., the server and the client.

\vspace*{0.03in}
- \textbf{Simulator for the corrupted server:}

\begin{adjustwidth}{20pt}{0pt}
a) \textbf{The case of intermediate layer.} $sim_{\mathcal{S}}$ 1) chooses an uniform random tape for the server; 2) sends model parameters $\bm{M}$ to $f^{\rm{ODT}}$ and gets the share of the nonlinear result as $out$;
2) randomly picks a public key $pk$ and encrypts all-zero input as $[0]_{sim_{\mathcal{S}}}$; 3) sends $[0]_{sim_{\mathcal{S}}}$ to server and receives the obscure linear result from the server; 4) encrypts $out$ with $\mathcal{S}$'s public key as $[out]_{\mathcal{S}}$; 5) sends $[out]_{\mathcal{S}}$ to server and outputs whatever $\mathcal{S}$ outputs. Here the view of server in real protocol execution is the client-encrypted input and the share of nonlinear function, while the simulated view is $sim_{\mathcal{S}}$-encrypted input and the same share of nonlinear function.
On the one hand, the client-encrypted input and $sim_{\mathcal{S}}$-encrypted input are indistinguishable due to the semantic security of HE. On the other hand, the share of nonlinear function are identical in real and simulated execution. So the output of
$sim_{\mathcal{S}}(\bm{M},out)$ is computationally indistinguishable to the $view_{\mathcal{S}}^\Pi$
of the corrupted server.
\end{adjustwidth}

\begin{adjustwidth}{20pt}{0pt}
b) \textbf{The case of last layer.} $sim_{\mathcal{S}}$ 1) chooses an uniform random tape for the server; 2) sends model parameters $\bm{M}$ to $f^{\rm{ODT}}$ and gets the None as $out$;
2) randomly picks a public key $pk$ and encrypts all-zero input as $[0]_{sim_{\mathcal{S}}}$; 3) sends $[0]_{sim_{\mathcal{S}}}$ to server and receives the obscure linear result from the server. Here the view of server in real protocol execution is the client-encrypted input while the simulated view is $sim_{\mathcal{S}}$-encrypted input.
As the client-encrypted input and $sim_{\mathcal{S}}$-encrypted input are indistinguishable due to the semantic security of HE, the output of
$sim_{\mathcal{S}}(\bm{M},out)$ is computationally indistinguishable to the $view_{\mathcal{S}}^\Pi$ of the corrupted server.
\end{adjustwidth}

\vspace*{0.03in}
- \textbf{Simulator for the corrupted client:}

\begin{adjustwidth}{20pt}{0pt}
a) \textbf{The case of intermediate layer.} $sim_{\mathcal{C}}$ 1) chooses an uniform random tape for the client; 2) sends private input $x$ to $f^{\rm{ODT}}$ and gets the share of the nonlinear result as $out$;
2) receives from client the $\mathcal{C}$-encrypted input as $[x]_{{\mathcal{C}}}$; 3) randomly forms a vector $r$ and encrypts it with client's public key as $[r]_{{\mathcal{C}}}$; 4) sends $[r]_{{\mathcal{C}}}$ to client and receives the $\mathcal{S}$-encrypted share of nonlinear function for server. Here the view of client in real protocol execution is the obscure linear result, e.g., $x'\circ{k'}\circ{v}+b$, while the simulated view is $r$.
As the $v$, $b$ and $r$ are random, $x'\circ{k'}\circ{v}+b$ and $r$ are indistinguishable. So the output of
$sim_{\mathcal{C}}(x,out)$ is computationally indistinguishable to the $view_{\mathcal{C}}^\Pi$
of the corrupted client.
\end{adjustwidth}

\begin{adjustwidth}{20pt}{0pt}
b) \textbf{The case of last layer.} $sim_{\mathcal{C}}$ 1) chooses an uniform random tape for the client; 2) sends private input $x$ to $f^{\rm{ODT}}$ and gets the obscure linear result as $out$;
3) receives from client the $\mathcal{C}$-encrypted input as $[x]_{{\mathcal{C}}}$; 3) enceypts $out$ with client's public key as $[out]_{{\mathcal{C}}}$; 4) sends $[out]_{{\mathcal{C}}}$ to client and outputs whatever $\mathcal{C}$ outputs. Here the view of client in real protocol execution and the simulated counterpart is identical. So the output of $sim_{\mathcal{C}}(x,out)$ are computationally indistinguishable to the $view_{\mathcal{C}}^\Pi$
of the corrupted client.
The proof of Theorem 1 is completed.
\end{adjustwidth}

}

\section{Performance Evaluation}\label{perf}
We implement CHEETAH with C++ based on Microsoft SEAL Library \cite{sealcrypto}, and compare it with the best existing scheme, GAZELLE\footnote{Available at: https://github.com/chiraag/gazelle\_mpc}.
We use two workstations as the client and server. Both machines run Ubuntu with Intel i7-8700 3.2GHz CPU with 12 threads and 16 GB RAM. The network link between them is a Gigabit Ethernet. Recall that the four parameters in BFV scheme are: 1) ciphertext modulus $q$; 2) plaintext modulus $p$; 3) number of ciphertext slots $n$ and 4) a Gaussian noise with a standard deviation $\sigma$. A larger $q/p$ tolerates more noise. We set $p$ to be a 20-bit number and $q$ to be a 60-bit psuedo-Mersenne prime. The number of slots for the packed encryption is set to 10,000.

\subsection{Component-wise Benchmark}
We first examine the performance of each functional component including Conv, FC and ReLu.

\noindent\textbf{Convolution Benchmark}. We define the time of the convolution operation as the duration between $\mathcal{S}$ receives the encrypted data or secret share from the previous layer (e.g., ReLu)  till $\mathcal{S}$ completes the convolution computation, just before sending the (partial) convolution results to $\mathcal{C}$. It does not contain the communication time between  $\mathcal{S}$ and $\mathcal{C}$, such as  transmitting the (partial) convolution results to $\mathcal{C}$, or secret share to $\mathcal{S}$, or in the case of GAZELLE, the time for the HE to GC transformation between $\mathcal{S}$ and $\mathcal{C}$ for fair comparison. All such communication time is accounted in ReLu and pooling discussed later.

Table \ref{convbench} benchmarks the convolution with different input and kernel sizes. The `In\_rot'  and `Out\_rot' indicate two GAZELLE variants with the input or output rotation, from which, one of them has to be used for convolution (see \cite{juvekar18gazelle} for details).
From Table \ref{convbench}, CHEETAH significantly outperforms GAZELLE. E.g., with the kernel size $5\times{5}@5$,  both the GAZELLE  In\_rot and Out\_rot  variants need more than 25 Mult, 24 Add and 24 Perm operations to yield the result of convolution. 
In contrast, CHEETAH needs only 5 Mult and 5 Add operations, one for each kernel,  to obtain the (partial) convolution results. Those results are then sent to $\mathcal{C}$ for computing ReLu (to be discussed).
Overall, CHEETAH accomplishes a speedup of $247$ and $207$ times compared with the GAZELLE In\_rot and Out\_rot variants, respectively, for the case with the kernel size $5\times{5}@5$ and input data size $28\times{28}@1$.


\begin{table}[tbp]
\centering
\footnotesize
\caption{Benchmark for convolution operation}
\vspace*{-0.15in}
\begin{tabular}{ccccc}
\hline
\hline
\begin{tabular}[c]{@{}c@{}}Input data size\\ ($w_i\times{h_i}@c_i$)\end{tabular} & \begin{tabular}[c]{@{}c@{}}Kernel size\\ ($k_p\times{k_q}@c_o$)\end{tabular} & Algorithm   & \begin{tabular}[c]{@{}c@{}}Time\\(ms)\end{tabular} &  Speedup\\
\hline
\hline
\multirow{3}{*}{28$\times${28}@1}  & \multirow{3}{*}{5$\times${5}@5} &  In\_rot& 7.4 & 247$\times$\\
 & & Out\_rot& 6.2 & 207$\times$\\
 & & \scriptsize{CHEETAH} & 0.03 & \\
\hline
\multirow{3}{*}{16$\times${16}@128} & \multirow{3}{*}{1$\times${1}@2} & In\_rot& 21.4 & 306$\times$\\
 & & Out\_rot& 4.65 & 66$\times$\\
 & &  \scriptsize{CHEETAH} & 0.07 & \multicolumn{1}{l}{}\\
\hline
\multirow{3}{*}{32$\times${32}@2} & \multirow{3}{*}{3$\times${3}@1} & In\_rot& 2.3& 115$\times$\\
 & & Out\_rot& 1.94 & 97$\times$\\
 & &  \scriptsize{CHEETAH} & 0.02 & \multicolumn{1}{l}{}\\
\hline
\hline
\end{tabular}
\label{convbench}
\vspace*{-0.1in}
\end{table}

\begin{figure}[!b]
\vspace*{-0.1in}
\centering
\includegraphics[scale=0.5]{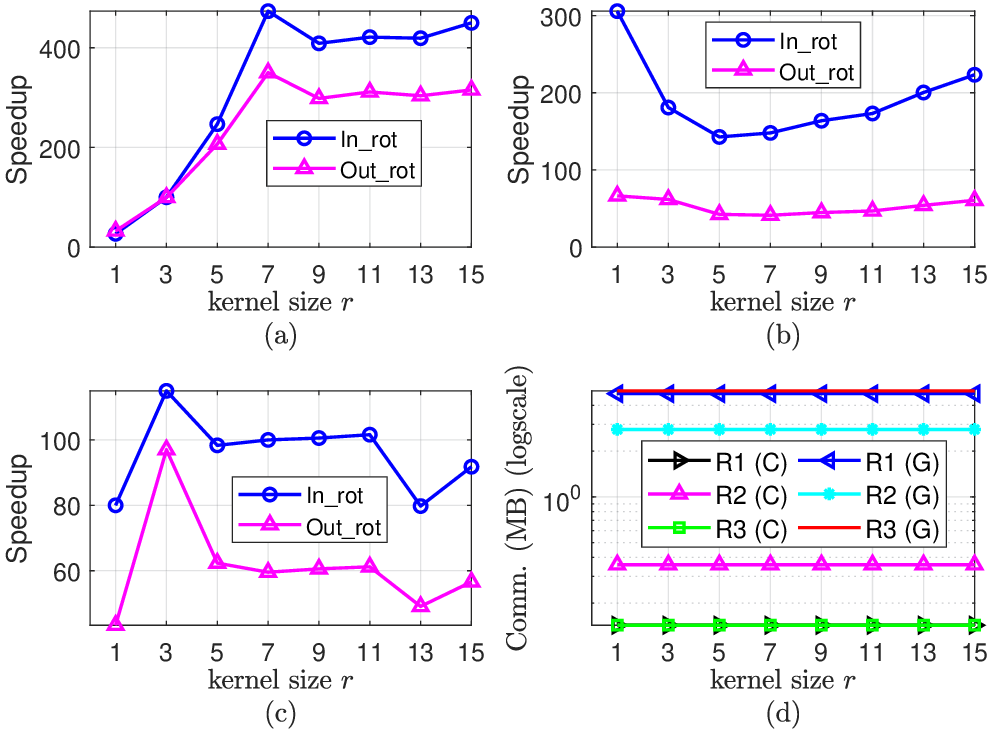}
\caption{Speedup and communication cost with various kernel sizes: (a) input data size 28$\times{28}@{1}$ and kernel size $r\times{r}@{5}$, (b) input data size 16$\times{16}@{128}$ and kernel size $r\times{r}@{2}$, (c) input data size 32$\times{32}@{2}$ and kernel size $r\times{r}@{1}$, (d) communication cost for a convolutional layer.}
\label{cvr}
\end{figure}

Fig. \ref{cvr} illustrates the speedup and communication cost vs. the kernel size $r$. Large kernel sizes have large receptive fields, thus capable of learning more information.\footnote{Classic structures such as VGG-16 cascade multiple small kernels to realize the same functionality of a large one.} CHEETAH offers more boost for large kernels, and achieves an average of $60$ (see Fig.~\ref{cvr}(c)) to $400$ (as shown in Fig.~\ref{cvr}(a)) speedup, with a slight exception when $r=1$. This is because with a larger kernel, GAZELLE conducts more expensive Permutations for the final convolutional results. When the kernel is sufficiently large, this speedup is offset by other computations hence the curve turns flat. This is reasonable since large kernels (e.g. $r>9$) are not desired in practice due to their heavy computations even in plaintext. Fig. \ref{cvr}(d) compares the communication cost (the size of the data sent to $\mathcal{C}$). For clarity, we sequentially denote the three rows of system configurations in Table \ref{convbench} as R1, R2 and R3. As can be seen, CHEETAH reduces the communication cost by $33$, $8$ and $35$ times for R1, R2 and R3, respectively.

\noindent\textbf{FC Benchmark}. Table \ref{matrivec} compares the  time for the weighted sum function in an FC layer (matrix-vector multiplication and nonlinear ReLu) for various input and output dimensions. The speedup of CHEETAH over GAZELLE is rather impressive, from about $300$ to over $400$ times, thanks to the cost savings from elimination of permutation. For example, when $n_o\times{n_i}$ is $1\times{2048}$, the input is a column vector with a size of 2048 which can be packed into one ciphertext. GAZELLE first conducts the Mult between the input and the weight matrix. As there are 2048 chunks with $n_o=1$, it then performs $\log_2(2048)=11$ Perm and $\log_2(2048)=11$ Add to compute the weighted sum. On the other hand, CHEETAH needs only one Mult and one Add, without the Perm. The results are packed into one ciphertext, sent to $\mathcal{C}$ and then used in the obscure HE to compute the nonlinear activation function. This gives CHEETAH $422$ times speedup compared with GAZELLE. From Table \ref{matrivec}, we can see that a larger ratio of $n_i/n_o$ leads to a higher speedup as GAZELLE  needs more Perm and Add operations. In contrast, \emph{CHEETAH needs only one Mult and one Add operation, independent of the $n_i/n_o$ ratio or $n_i$, $n_o$ value}. This is more beneficial for designing privacy-preserved, large-scale learning tasks since the objective is to map the inputs into a high-dimensional vector in order to characterize all the classes. The large $n_i/n_o$ ratio is common in networks such as VGG16, GoogleNet and ResNet50 for the ImageNet task.

\begin{table}[tbp]
\centering
\footnotesize
\caption{Benchmark for matrix-vector mult.}
\vspace*{-0.13in}
\begin{tabular}{ccC{0.55cm}C{0.55cm}C{0.55cm}cc}
\hline
\hline
$n_o\times{n_i}$ & Method & \#Perm & \#Mult & \#Add & \begin{tabular}[c]{@{}c@{}}Time\\(ms)\end{tabular} & Speedup \\
\hline
\hline
\multirow{2}{*}{1$\times$2048} & \scriptsize{GAZELLE} & 11 & 1 & 11 & 3.8 & 422$\times$ \\
 &  \scriptsize{CHEETAH} & 0 & 1 & 1 & 0.009 & \\
\hline
\multirow{2}{*}{2$\times$1024} & \scriptsize{GAZELLE} & 10 & 1 & 10 & 3.63 & 403$\times$ \\
 &  \scriptsize{CHEETAH} & 0 & 1 & 1 & 0.009 & \\
\hline
\multirow{2}{*}{4$\times$512} & \scriptsize{GAZELLE} & 9 & 1 & 9 & 3.3 & 367$\times$ \\
 &  \scriptsize{CHEETAH} & 0 & 1 & 1 & 0.009 & \\
\hline
\multirow{2}{*}{8$\times$256} & \scriptsize{GAZELLE} & 8 & 1 & 8 & 3 & 333$\times$ \\
 & \scriptsize{CHEETAH} & 0 & 1 & 1 & 0.009 & \\
\hline
\multirow{2}{*}{16$\times$128} & \scriptsize{GAZELLE} & 7 & 1 & 7 & 2.65 & 294$\times$ \\
 & \scriptsize{CHEETAH} & 0 & 1 & 1 & 0.009 & \\
\hline
\hline
 \end{tabular}
 \label{matrivec}
\vspace*{-0.05in}
\end{table}


Table~\ref{FCcomm} presents the communication cost for the FC layer. As can be seen, GAZELLE has a higher communication cost due to GC, especially for a large output dimension $n_o$, while \emph{the  communication cost of CHEETAH is independent of the input or output dimensions}, as it needs only one ciphertext. Using GC, GAZELLE needs more communication overhead between $\mathcal{S}$ and $\mathcal{C}$.
Note that this is also true for the communication time after convolution, for computing the nonlinear ReLu activation.

\begin{table}[tbp]
\centering
\footnotesize
\caption{Commun. cost for matrix-vector mult. (KB)}
\vspace*{-0.16in}
\begin{tabular}{cccccc}
\hline\hline
$n_o\times{n_i}$ & 1$\times$2048 & 2$\times$1024 & 4$\times$512 & 8$\times$256 & 16$\times$128 \\ \hline
\hline
CHEETAH & 143.1  &  143.1  &   143.1   &  143.1  &  143.1    \\ \hline
GAZELLE & 147.8   &  152.5  &   161.9   &  180.8   &  218.6   \\ \hline\hline
 \end{tabular}
 \label{FCcomm}
\vspace*{-0.1in}
\end{table}

\noindent\textbf{ReLu Benchmark}. Table \ref{relumax} shows the speedup of nonlinear operations, i.e., the ReLu function. As its obscure HE only conducts $0$-multiplicative-depth HE operation to compute the ReLu function, followed by a one-way communication from $\mathcal{C}$ to $\mathcal{S}$,  to send the ReLu share, CHEETAH dramatically improves the efficiency of  nonlinear operations by up to $1793$ times for ReLu, compared with GAZELLE.
\begin{table}[tbp]
\centering
\footnotesize
\caption{Benchmark for ReLu operation.}
\vspace*{-0.18in}
\begin{tabular}{ccccc}
\hline
\hline
Function & \begin{tabular}[c]{@{}c@{}}Output\\ dimension\end{tabular} & Method & \begin{tabular}[c]{@{}c@{}}Online\\ cost (ms)\end{tabular} & Speedup \\
\hline
\hline
\multirow{4}{*}{\textit{ReLu}} & \multirow{2}{*}{1000} & \scriptsize{GAZELLE} & 115 & 267$\times$ \\
 &  & \scriptsize{CHEETAH} & 0.39 &  \\
\cline{2-5}
 & \multirow{2}{*}{10000} & \scriptsize{GAZELLE} & 843 & 1793$\times$ \\
 &  & \scriptsize{CHEETAH} & 0.47 &  \\
\hline
\hline
\end{tabular}
\label{relumax}
\end{table}

\begin{figure}[!tbp]
\centering
\includegraphics[scale=0.5]{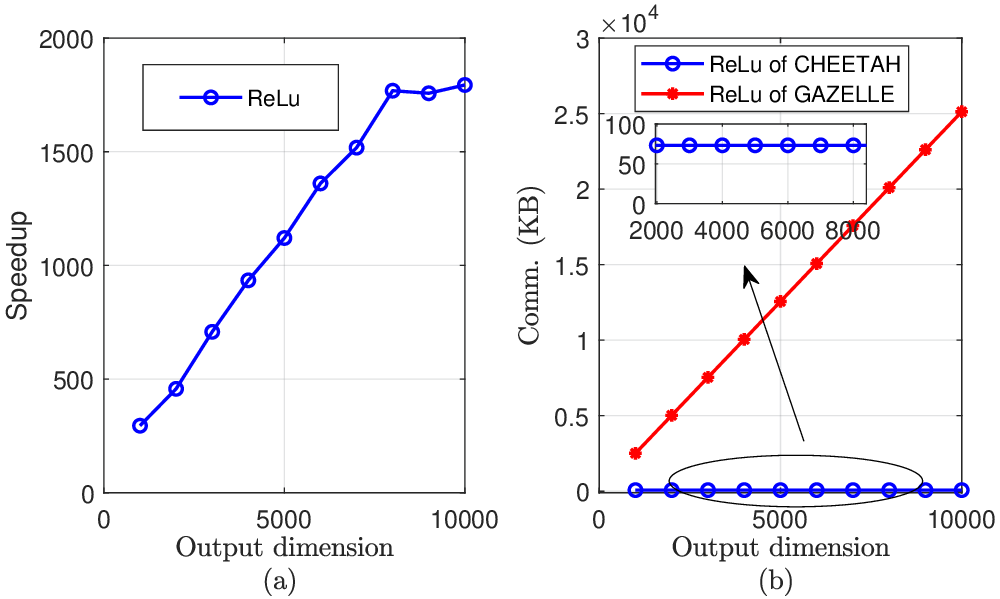}
\vspace*{-0.1in}
\caption{(a) Speedup of CHEETAH over GAZELLE for computing ReLu. (b) Comparison of communication cost for ReLu. }
\label{nonli}
\end{figure}

Fig. \ref{nonli} plots the  speedup and communication cost as a function of the  output dimension. Similarly, CHEETAH achieves an outstanding speedup with much smaller communication cost, independent of the output dimension, compared with GAZELLE. The speedup quickly  increases  when the output dimension increases. The communication cost of CHEETAH only involves the number of packed ciphertexts for nonlinear share of $\mathcal{S}$. CHEETAH needs only one round of communications.  In comparison, GAZELLE needs the  GC module to obtain the nonlinear result, which has a large communication cost proportional to the output dimension, and needs multiple rounds of communications between $\mathcal{C}$ and $\mathcal{S}$.
Overall, CHEETAH achieves a communication cost reduction up to \emph{two orders of magnitude} compared with GAZELLE.

\begin{figure}[!t]
\centering
\includegraphics[trim= {1.3cm 0.4cm 1.92cm 0.5cm}, clip, scale=0.425]{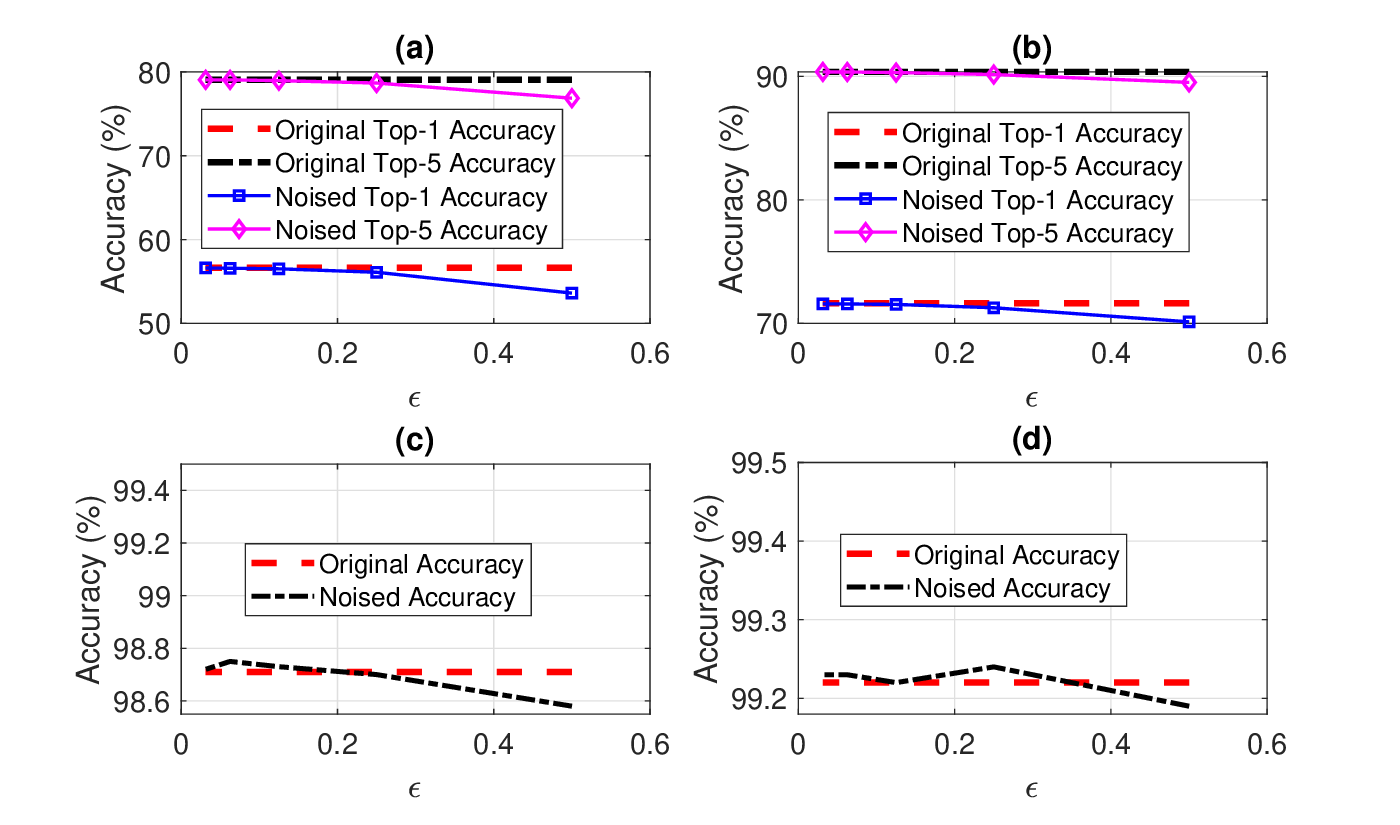}
\vspace*{-0.1in}
\caption{\color{black}Accuracy with different noise range for: (a) AlexNet; (b) VGG-16; (c) Network A; and (d) Network B.}
\label{noised_accu}
\end{figure}

\begin{figure*}[!htbp]
\centering
\includegraphics[scale=0.82]{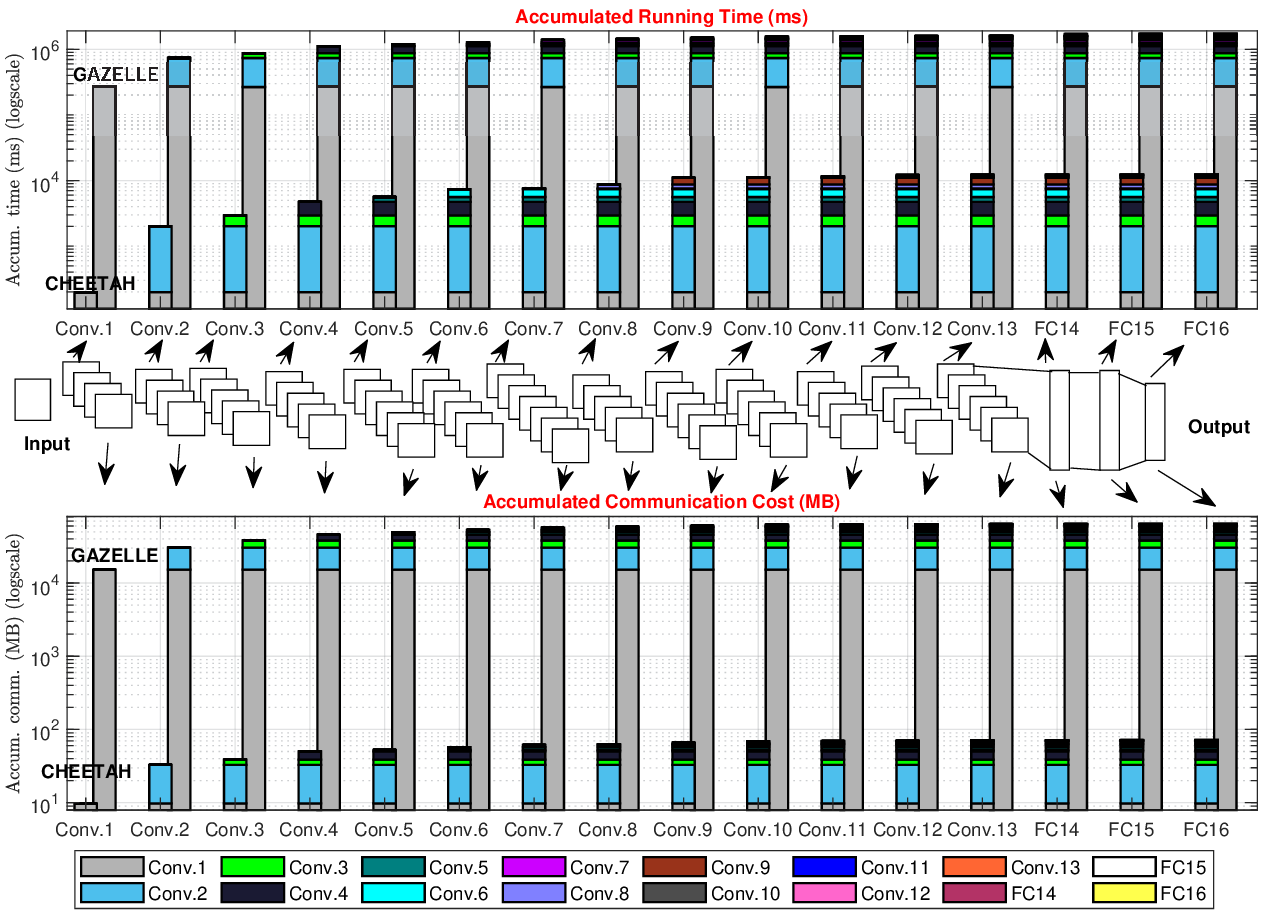}
\caption{Benchmark for VGG-16. (best view in color)}
\label{VGG16}
\end{figure*}

\subsection{Benchmark with Classic Networks}
This section compares the overall performance of CHEETAH and GAZELLE in complete networks from end to end, i.e., including all the computation and communication from the data input to the final inference result. We benchmark on four neural network structures: (i) \textbf{Network A}~\cite{rouhani2018deepsecure}: 1 Conv  and 2 FC layers with ReLu activation, (ii) \textbf{Network B}~\cite{liu2017oblivious}: 2 Conv  and 2 FC layers with ReLu activation and pooling, (iii) \textbf{AlexNet}~\cite{krizhevsky2012imagenet}: $5$ Conv  and $3$ FC layers with ReLu activation and pooling, (iv) \textbf{VGG-16}~\cite{simonyan2014very}: $13$ Conv  and $3$ FC layers with ReLu activation and pooling. Networks A and B represent relatively shallow structures like LeNet~\cite{lecun1998gradient} that were used in simple tasks like recognition of handwritten digits. AlexNet and VGG-16 achieve record-breaking performance on large-scale dataset like ImageNet~\cite{krizhevsky2012imagenet}. These networks leverage a stack of convolutional layers to extract complex feature relations from inputs of large dimensions ($224\times224\times3$ RGB images). Thus, the capability of running such deep networks in reasonable time is a significant contribution to bridge the gap from research to practical privacy-preserved computer vision applications.

Table \ref{prenet} presents the  speedup of CHEETAH over GAZELLE.  CHEETAH achieves $130$ to $334$ times speedup across the four networks. It not only dramatically reduces the running time compared with GAZELLE, but also brings the running time down to the \emph{practical} range. For instance, for the VGG-16 network, GAZELLE needs about half an hour to get an inference result (image classification), while CHEETAH takes only 12 seconds, which is practical for many MLaaS applications, or even potential mobile applications.  \emph{This is the first time that the privacy preserved learning can be applied to  practical neural networks for real world applications}. Table \ref{prenet}  also depicts the communication cost of both schemes in the unit of MB. CHEETAH accomplishes from $8.9$ to $38$ times communication cost reduction, compared with GAZELLE. This is a critical improvement that not only relieves the requirement for high bandwidth but also reduces the communication delay.

{\color{black}Meanwhile, as shown in Sec.~\ref{sec.3.1}, CHEETAH introduces a noise term $\delta_i$ that is uniformly distributed between $[-\epsilon,\epsilon]$ for each linear output in each layer and Fig~\ref{noised_accu} shows the relation between $\epsilon$ and system accuracy of the neural networks. We can see that the system accuracy is kept with negligible drop for each networks as $\epsilon<0.25$.
Therefore, $\epsilon$ (together with $v$ in Sec.~\ref{sec.3.1}) can be secretly chosen by server to hide the linear result as well as keep the system performance.}

Next we examine the running time breakdown at each layer. This would be quite helpful to locate the performance bottleneck in the network. Fig. \ref{VGG16} shows the accumulated running time and communication cost of VGG-16 network from the first to the last layer. Each bar accumulates the running time or communication cost breakdown from the first layer to the current layer, e.g., the bars of ``Conv.2'' denote the running time or communication cost up till Conv.2. For better visualization of the  speedup, the time-axis is plotted in log scale.

As can be seen, the layers at the beginning of the network give great speedup. This is due to the nature of deep neural networks as the dimensionality of data is reduced by convolution and pooling layers towards the end of the network. Recall that CHEETAH enjoys a large acceleration by avoiding the Perm operations, especially when the data dimensions are large. Thus, it can easily handle large images with high resolutions, and alleviates the speed bottleneck at the beginning of those deep networks.



%

\begin{table}[t]
\centering
{\scriptsize
\caption{Benchmark for classic networks.}
\begin{tabular}{cccccc}
\hline
\hline
Metrics &Framework & Net. A & Net. B & AlexNet & VGG-16\\
\hline\hline
\multicolumn{6}{c}{Online}\\
\hline
\multirow{3}{*}{Time(ms)} & GAZELLE & 190 & 1689 & 161,509 & 1,731,151\\
\cline{2-6}
 & CHEETAH & 0.87 & 5.05&1,247 & 12,338\\
\cline{2-6}
 & Speedup & 218$\times$ & 334$\times$ & 130$\times$ & 140$\times$\\
\hline
\multirow{3}{*}{Comm.(MB)}&GAZELLE& 5.531& 118.8 & 3113 & 64197 \\
\cline{2-6}
 & CHEETAH & 0.429 & 0.643& 9.15 & 71.9 \\
\cline{2-6}
 & Reduction & 13$\times$ & 185$\times$ & 340$\times$ & 893$\times$ \\
\hline
\hline
\multicolumn{6}{c}{Offline}\\
\hline
\multirow{3}{*}{Time(ms)} & GAZELLE & 4 & 42.3 & 2,796.3 & 55,524.2\\
\cline{2-6}
 & CHEETAH & 0.2  & 3.5 & 63.6 & 1,388 \\
\cline{2-6}
 & Speedup &  20$\times$ &  12$\times$ &  44$\times$ &  40$\times$\\
\hline
\multirow{3}{*}{Comm.(MB)}&GAZELLE& 4.6& 50.4 & 3214.2 & 66190 \\
\cline{2-6}
 & CHEETAH & 0.21 & 4.1& 72.1 & 1640 \\
\cline{2-6}
 & Reduction & 21$\times$ & 12.3$\times$ & 44.6$\times$ & 40.4$\times$ \\
\hline
\hline
\end{tabular}
\label{prenet}
\vspace*{-0.13in}
}
\end{table}

\section{Conclusion and Future Work}\label{conclu}
In this work we have focused on ensuring privacy-preserved inference in Machine Learning as a Service (MLaaS). In particular, we have unveiled the fundamental performance bottleneck in  existing state-of-the-art schemes, and proposed CHEETAH, an ultra-fast, secure MLaaS framework, that features a carefully crafted secret sharing scheme to enable efficient, joint linear and nonlinear computation, to make CHEETAH run significantly faster than the state-of-the-art solutions. CHEETAH  eliminates the need to use approximation for nonlinear functions, and hence does not have accuracy loss, unlike the existing schems. It has been evaluated on the benchmark of well-known, practical deep networks such as AlexNet and VGG-16 on the MNIST and ImageNet datasets.  The results have demonstrated  218 and 334 times  speedup over  the state-of-the-art GAZELLE for a 3-layer and a 4-layer CNN, respectively, and 130 and 140 speedup over GAZELLE in  AlexNet and VGG-16, respectively. Compared with CryptoNets, CHEETAH achieves a speedup of five orders of magnitudes. By our best knowledge, this is the first time to demonstrate that privacy-preserved deep neural networks can be applied in real world applications.


\bibliographystyle{IEEEtran}
\bibliography{ccs-sample}
\vspace*{-0.7in}
\begin{IEEEbiography}[{\includegraphics[trim= {0.6cm 0.1cm 0.6cm 0cm}, clip, scale=0.13]{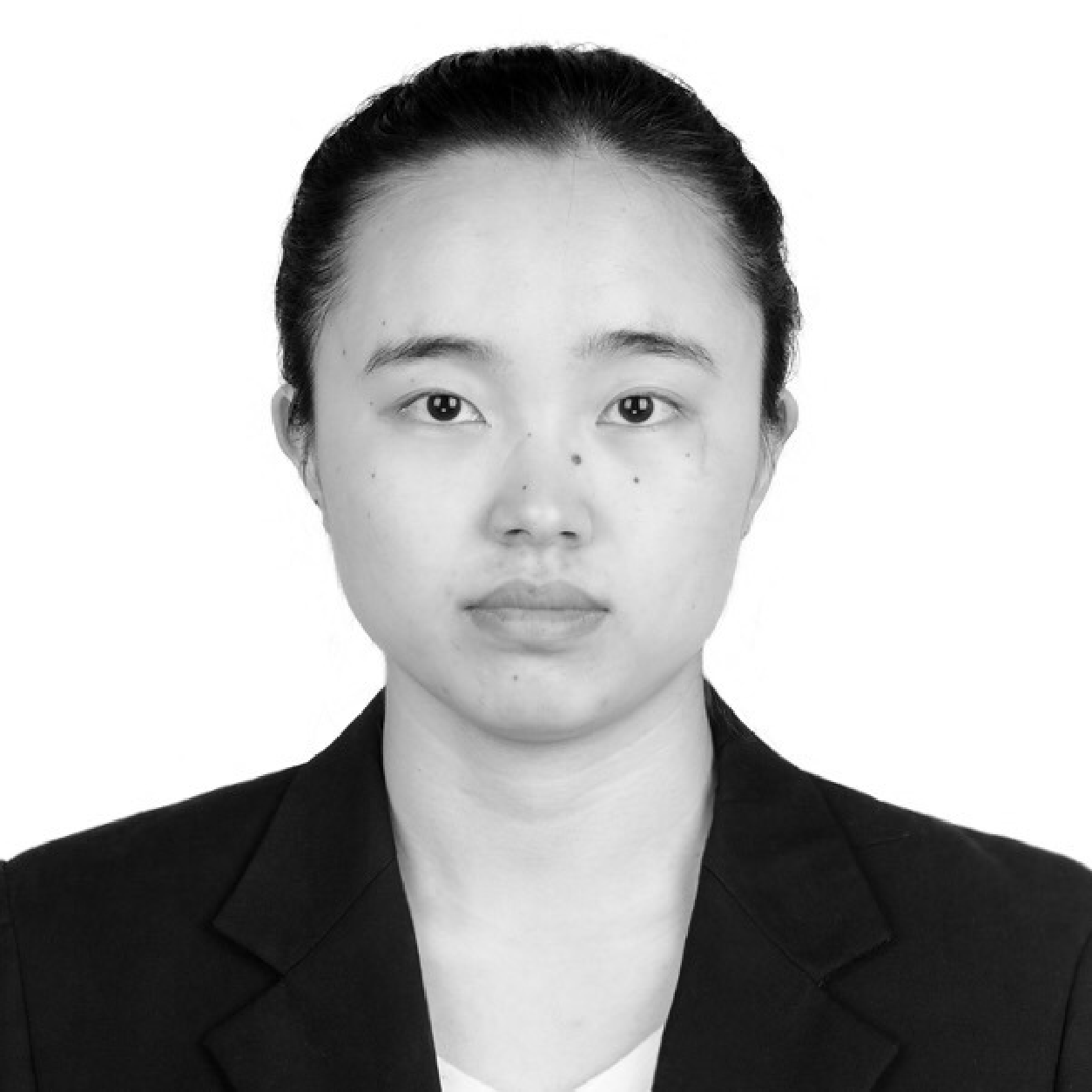}}]{Qiao Zhang}
received the B.S. (14') and M.S. (17') degrees from the School of Communication and Information Engineering, Chongqing University of Posts and Telecommunications, Chongqing, China. She is currently a Ph.D. candidate with the Department of Electrical and Computer Engineering, Old Dominion University (ODU), Norfolk, VA, USA. Her current research focuses on privacy preserving machine learning.
\end{IEEEbiography}
\vspace*{-0.9in}
\begin{IEEEbiography}[{\includegraphics[trim= {0.6cm 0.1cm 0.6cm 0cm}, clip, scale=0.065]{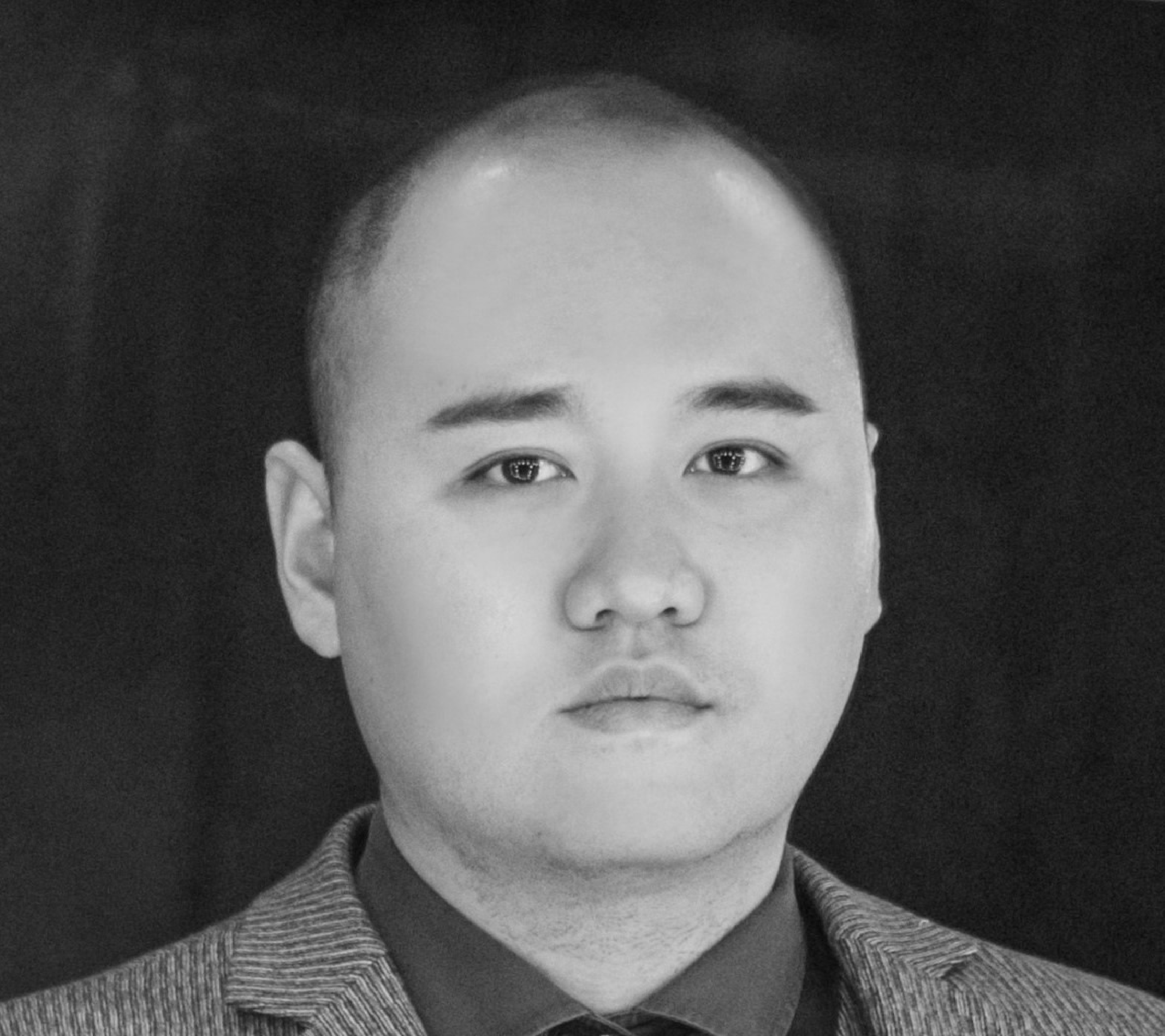}}]
{Cong Wang} received the B. Eng degree in Information Engineering from the Chinese University of Hong Kong in 2008, M.S. degree in Electrical
Engineering from Columbia University in 2009,
and Ph.D. in Computer and Electrical Engineering
from at Stony Brook University, NY, in 2016. He is
currently an Assistant Professor at the Computer
Science department, Old Dominion University,
Norfolk, VA. His research focuses on exploring
algorithmic solutions to address security and
privacy challenges in Mobile, Cloud Computing,
IoT, Machine Learning and System. He is the recipient of Commonwealth
Cyber Initiative Research and Innovation Award, ODU Richard Cheng
Innovative Research Award and IEEE PERCOM Mark Weiser Best Paper
Award in 2018.
\end{IEEEbiography}
\vspace*{-2.75in}
\begin{IEEEbiography}[{\includegraphics[trim= {0.6cm 0.1cm 0.6cm 0cm}, clip, scale=0.033]{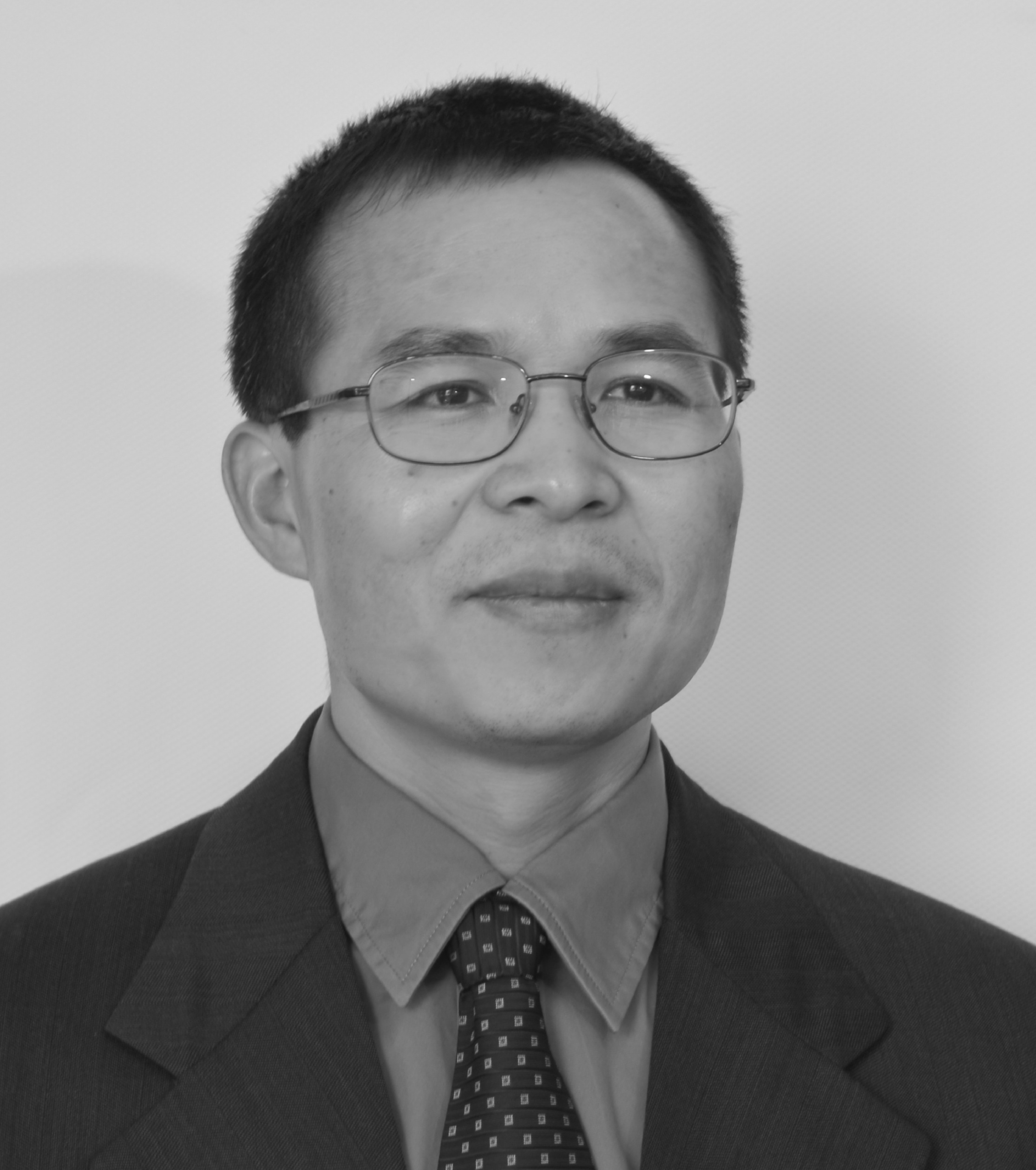}}]{Chunsheng Xin}
is a Professor in the Center for Cybersecurity Education and Research, and the Department of Electrical and Computer Engineering, Old Dominion University. He received his Ph.D. in Computer Science and Engineering from the State University of New York at Buffalo in 2002. His interests include cybersecurity, machine learning, wireless communications and networking, cyber-physical systems, and Internet of Things. His research has been supported by almost 20 NSF and other federal grants, and results in more than 100 papers in leading journals and conferences, including three Best Paper Awards, as well as books, book chapters, and patent. He has served as Co-Editor-in-Chief/Associate Editors of multiple international journals, and symposium/track chairs of multiple international conferences including IEEE Globecom and ICCCN. He is a senior member of IEEE.
\end{IEEEbiography}
\vspace*{-2.75in}
\begin{IEEEbiography}[{\includegraphics[trim= {0.6cm 0.1cm 0.6cm 0cm}, clip, scale=0.075]{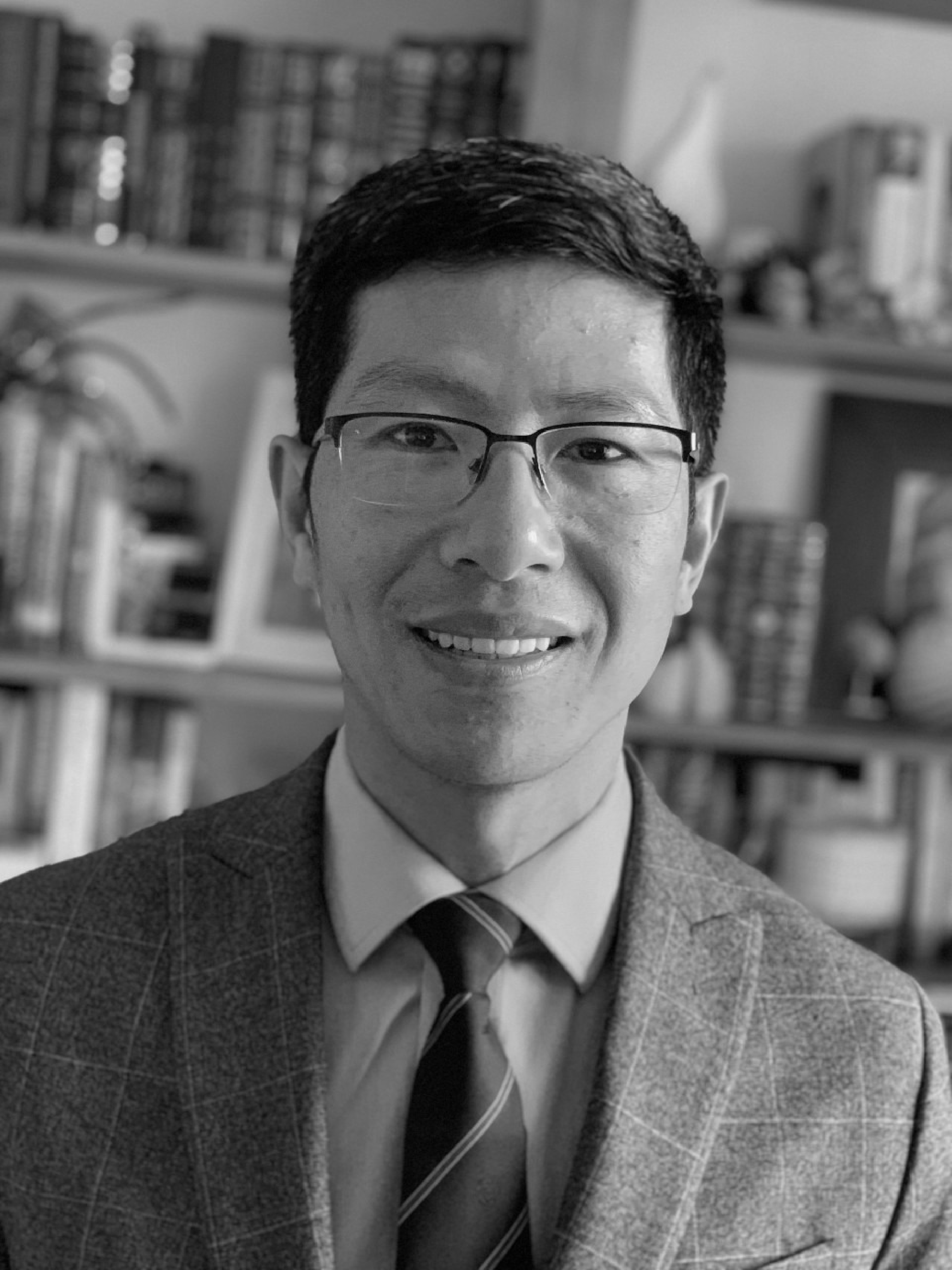}}]{Hongyi Wu}
is the Batten Chair of Cybersecurity and the Director of the Center for Cybersecurity
Education and Research at Old Dominion University (ODU). He is also a Professor in Department of Electrical and Computer Engineering and holds joint appointment in Department of
Computer Science. Before joining ODU, he was an Alfred and Helen Lamson Endowed Professor at the Center for Advanced Computer Studies (CACS), University of Louisiana at Lafayette
(UL Lafayette). He received the B.S. degree in scientific instruments from Zhejiang University, Hangzhou, China, in 1996, and the M.S. degree in electrical engineering and Ph.D. degree
in computer science from the State University of New York (SUNY) at Buffalo in 2000 and 2002, respectively. His research focuses on networked cyber-physical systems for security, safety, and emergency management applications, where the devices are often light-weight, with extremely limited computing power, storage space, communication bandwidth, and battery supply. He received NSF CAREER Award in 2004 and UL Lafayette Distinguished Professor Award in 2011. He is a fellow of IEEE.
\end{IEEEbiography}


%

%
%
%




\end{document}